**Front Matter**

**Title**

Divergent Creativity in Humans and Large Language Models

Short Title: Divergent Creativity in Humans and LLMs

**Authors**

Antoine Bellemare-Pepin[1,2,†], François Lespinasse[3,†], Philipp Thölke[1], Yann Harel[1], Kory Mathewson[4], Jay A. Olson[5], Yoshua Bengio[4,6] and Karim Jerbi[1,4,7,*]

**Affiliations**

[1] CoCo Lab, Psychology department, Université de Montréal, Montreal, QC, Canada

[2] Music department, Concordia University, Montreal, QC, Canada

[3] Sociology and Anthropology department, Concordia University, Montreal, QC, Canada

[4] Mila (Quebec AI research Institute), Montreal, QC, Canada

[5] Department of Psychology, University of Toronto Mississauga, Mississauga, ON, Canada

[6] Department of Computer Science and Operations Research, Université de Montréal, Montreal, QC, Canada

[7] UNIQUE Center (Quebec Neuro-AI research Center), QC, Canada

* : Corresponding author : Karim Jerbi; karim.jerbi@umontreal.ca

† : These authors contributed equally to this work.

## Abstract

The recent surge of Large Language Models (LLMs) has led to claims that they are approaching a level of creativity akin to human capabilities. This idea has sparked a blend of excitement and apprehension. However, a critical piece that has been missing in this discourse is a systematic evaluation of LLMs' semantic diversity, particularly in comparison to human divergent thinking. To bridge this gap, we leverage recent advances in computational creativity to analyze semantic divergence in both state-of-the-art LLMs and a substantial dataset of 100,000 humans. We found evidence that LLMs can surpass average human performance on the Divergent Association Task, and approach human creative writing abilities, though they fall short of the typical performance of highly creative humans. Notably, even the top performing LLMs are still largely surpassed by highly creative individuals, underscoring a ceiling that current LLMs still fail to surpass. Our human-machine benchmarking framework addresses the polemic surrounding the imminent replacement of human creative labour by AI, disentangling the quality of the respective creative linguistic outputs using established objective measures. While prompting deeper exploration of the distinctive elements of human inventive thought compared to those of AI systems, we lay out a series of techniques to improve their outputs with respect to semantic diversity, such as prompt design and hyper-parameter tuning.

## Significance statement

The rise of generative AI systems has led to provocative claims that they have reached human-level creativity. We probe this question through an unprecedented large-scale assessment of creativity in multiple Large Language Models (LLMs) and in a large human dataset of 100,000 individuals. Our findings show that in the divergent association task LLMs can surpass the average performance of individuals from the general population but not that of highly creative individuals. Additionally, we show that LLMs can approach human-level performance in creative writing, based on automated scoring methods with known limitations. Critically, we also provide an assessment of the impact of LLM hyperparameter tuning on its creativity-related measures. The quantitative benchmarking framework introduced by this study marks a significant step in the field of computational creativity and opens new paths for the assessment and development of AI creativity.

**MAIN TEXT**

## Introduction

Creativity is a multifaceted construct at the crossroads of individual expression, problem solving, and innovation. Human creativity is pivotal in shaping cultures and has undergone continuous transformation across historical epochs. Our understanding of this ability is now influencing the landscape of artificial intelligence and cognitive systems (*1*–*5*). In the past few years, the advent of sophisticated Large Language Models (LLMs) has spurred considerable interest in evaluating their capabilities and apparent human-like traits (*6*), particularly in terms of their impacts on human creative processes (*7*, *8*). Despite a growing interest in evaluating the creative quality of LLM-generated outputs (*9*–*12*), current benchmarking approaches have yet to systematically compare LLMs to human performance on tasks that are suitable for both.

Although the ability to generate novel and aesthetically pleasing artifacts has long been considered a uniquely human attribute, this view has been challenged by the recent advances in generative AI. This technological progress has ignited discussions surrounding the creative capabilities of machines (*13*–*16*), ushering in the emerging field of computational creativity—a multidisciplinary domain that explores the potential of artificial systems to exhibit creativity in a manner analogous to human cognition.

The release of GPT-4 was marked with an exceptional gain in performance across various standardized benchmarks (*17*). Demonstrating its versatility in language- and vision-based tasks, GPT-4 has successfully passed a uniform bar examination, the SAT, and multiple AP exams, transcending the boundaries of traditional AI capabilities. However, it is important to keep in mind that such benchmarks can be achieved through non-human processes such as data contamination and storage, rather than genuine reasoning or understanding. The model's web page (openai.com/gpt-4) touts its creative prowess, spurring a fresh examination of the creativity of state-of-the-art LLMs. The stance taken by OpenAI has sparked debates on the extent to which the creativity of LLMs is poised to rival human capabilities.

These advancements raise pivotal questions for the science of creativity: Are these models genuinely evolving to become more creative, and to what extent do they approach human-level creativity? The exploration of these inquiries not only deepens our understanding of artificial creativity but also provides valuable insights into the role that language abilities play in creativity.

Here, we leverage recent computational advances in the field of creativity science in order to quantify creativity across state-of-the-art LLMs and in a massive data set of 100,000 human participants. By scrutinizing these models through the lens of distributional semantics, we probe and compare their potential to generate original linguistic and narrative content.

Numerous definitions and frameworks have been proposed to describe human creativity, encompassing convergent and divergent thinking, as well as variation-selection paradigms (*2*, *8*, *15*, *18*–*20*). Divergent thinking, characterized by the ability to generate novel and diverse solutions to open-ended problems, has gained widespread recognition as a robust and widely-accepted index of creative cognition (*21*). This aspect of cognitive creativity is particularly tied to the initial phase of the creative process (i.e., variation/exploration), where many ideas are produced before the most useful and novel ones are selected.

To quantify divergent thinking, researchers have employed various tools, such as the Alternative Uses Test (AUT), in which people generate novel uses for common objects. Recently, the creativity of LLMs has been probed using the AUT, yielding mixed results; while there were no overall significant differences between LLMs and humans, discrepancies emerged in specific AUT items (*22*, *23*). The results might be explained by inherent challenges in the methodology (*24*). The AUT's validity remains contentious (*25*), and chatbot responses might inadvertently draw from online test materials. Additionally, their methodology of eliciting multiple responses from chatbots has raised concerns over the significance of fluency metrics. This aligns with broader critiques of the AUT, highlighting its cumbersome and subjective rating process (*26*), even if recent work has shown promising approaches using LLMs to automatically score the AUT (*27*).

More recently, semantic distance is increasingly probed as a key component of creative thought (*28*). Recent methodological advances include, for instance, the Divergent Association Task (DAT), in which people are asked to generate a list of 10 words that are as semantically distant from one another as possible (*29*). Individuals who are more creative tend to cover a larger semantic repertoire, resulting in a larger mean semantic distance between the words. While the DAT predominantly probes the *novelty* aspect of creativity through language, its results have been found to correlate with other established creativity tests such as the AUT, thus validating its reliability as a measure of creative potential in humans (*29*–*33*).

The speed and unambiguous scoring of the DAT make it appropriate for large-scale evaluations. It may be useful to assess both LLMs and human creativity, as it is a straightforward task that probes creative potential through language production, a domain accessible to both entities. This commonality facilitates a concise and direct comparison of creative output between LLM models and humans, enabling an in-depth examination of their respective creative capacities. Further, the DAT uses computational scoring to assess semantic distance between all word pairs, allowing the comparison of large samples without additional bias from human raters. Semantic distance is derived from the mean cosine similarity value between pairs of word embeddings—matrix-based representations of words. These embeddings are produced by a language model that is trained to consider word co-occurrences, a characteristic often termed as context-independent word embeddings (*34*).

An alternative method for evaluating creativity is through the examination of creative writing. Recent investigations have used a quantitative approach similar to that taken by the DAT to assess the semantic distance covered by sentence-based texts (*35*). Divergent Semantic Integration (DSI) is a measure of cosine similarity between pairs of word-level embeddings present in a textual narrative. This approach was implemented in light of more recent advances in language modeling allowing the computation of context-dependent word embeddings, which take the entire surrounding sentence into account (*36*). DSI has been found to correlate strongly with human ratings of perceived creativity in short narratives (*35*).

The research community has recently delved into investigating the creative behavior of LLMs (*7*, *22*, *37–43*) and exploring the potential interactions between human and machine creativity (*24*, *44–49*). Recent studies have further advanced this field by evaluating creative writing in LLMs from diverse perspectives—comparing GPT-4 to award-winning novelist Patricio Pron in a human–machine creative writing contest (*50*), demonstrating that LLM productions can match human-level creativity on certain humor and epicness dimensions (*51*), and introducing novel automated methods for analyzing story arcs, turning points, and affective dynamics (*52*)—which we complement by directly comparing both DAT scores and performance on diverse creative writing tasks. However, a comprehensive benchmark analysis comparing creativity, measured by semantic divergence, across state-of-the-art LLMs and human performance is lacking. Our study not only seeks to fill this gap empirically but also to discuss the potential implications of applying creativity measures to AI productions on our understanding of human cognition and creative potential.

This paper provides a thorough examination of the ability of LLMs to mimic human creativity by comparing each others' performance using established creativity measurements. To this end, we collected responses on the DAT from a large cohort of 100,000 participants and compared them with the performance of a diverse array of LLMs. We further explored the influence of hyperparameter tuning and prompting strategies. Furthermore, we tested the hypothesis that models exhibiting higher performance on the DAT would similarly excel in a set of creative writing tasks, as compared to human-generated content.

The LLMs assessed in this study were not selected with the intent of conducting a comprehensive and competitive comparison of the best models available. The sheer pace of current LLM development would render such an approach quickly obsolete. Instead, we chose a wide range of models that vary in characteristics such as size, popularity, training, and license, hoping to provide a general framework to assess creativity in LLMs as compared to human participants. Throughout the manuscript, we use the term 'LLM creativity' to refer specifically to the divergent aspect of semantic creativity, i.e. the ability to produce highly dissimilar sets of words, or in the case of story-writing, to integrate diverse ideas, objects, etc. into a narrative.. As demonstrated by previous research using the DAT and DSI, this dimension of creativity shows a strong correlation with other facets of creative processes in humans (*29*, *35*). Coherently, our evaluations do not presume that LLMs achieve similar performance using similar processes, and rather present a framework for Human-AI benchmarking on these tests, which may eventually lead to a more granular analysis of the implicated processes.

# Results

## Comparing Large Language Models (LLMs) and human creativity using the Divergent Association Task

To benchmark the divergent creativity of humans and different LLMs, we compared the mean of their respective DAT scores *(see Methods)*. As depicted in **Figure 1A**, GPT-4 surpasses human scores with a statistically significant margin, followed by GeminiPro, which is statistically indistinguishable from human performance. Interestingly, Vicuna, a drastically smaller model, performs significantly better than some of its larger counterparts. Apart from the Humans/GeminiPro, GeminiPro/Claude3 and Vicuna/GPT-3.5 contrasts, all other pairwise contrasts of mean DAT score are statistically significant (**Fig. 1B**). Importantly, a later release from OpenAI, GPT-4-turbo, demonstrates a notable decline in performance when compared to its predecessor, GPT-4. A comprehensive analysis across all versions of the GPT-4 models, as illustrated in **Figure S2**, indicates that newer iterations of the model do not consistently enhance performance on the DAT.

Notably, models with lower scores exhibit greater variability (**Fig. 1C**), often coinciding with a greater tendency to fail to comply with the instruction (as depicted by the pie charts).

The word count analysis (see **Fig. 1D**) revealed that GPT-4-turbo showed the highest degree of word repetition across all responses with the word *ocean* occurring in more than 90% of the word sets. The best performing model, GPT-4, also showed a high degree of word repetition across all responses with 70% of responses containing the word *microscope*, followed by *elephant* (60%). The latter was ranked first in GPT-3.5's responses, while the most frequent words chosen by humans were *car* (1.4%) followed by *dog* (1.2%) and *tree* (1.0%).

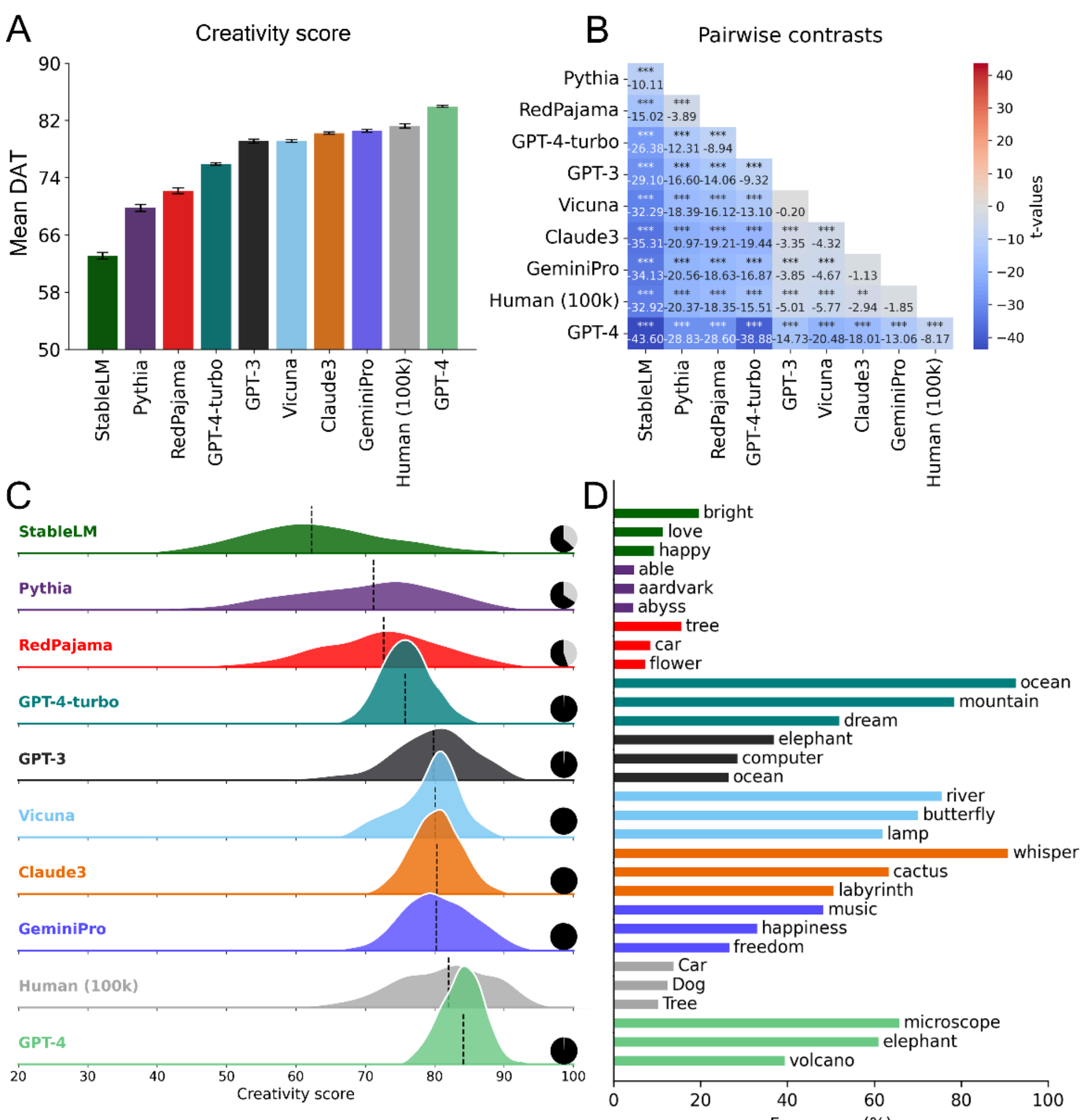

**Fig 1. Comparing LLMs and humans on the Divergent Association Task (DAT).** Summary of DAT performance across LLM and human samples. **(A)** Mean DAT score and 95% confidence intervals. **(B)** Heatmap of all contrasts, generated using two-sided independent t-tests, sorted by their correlation with the highest performing model, GPT-4. **(C)** Distribution for each model using a ridge plot of smoothed kernel density estimates. Black vertical lines represent the mean, and the small black/gray pie charts show the models' prompt adherence (i.e. the proportion of valid responses). **(D)** Most frequent words across responses. The percentages represent the proportion of response sets (10 words) that include these words. *: p<.05, **: p<.01, ***: p<.001.

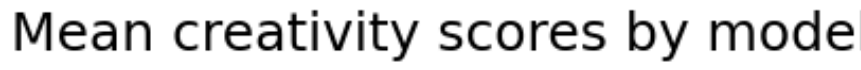

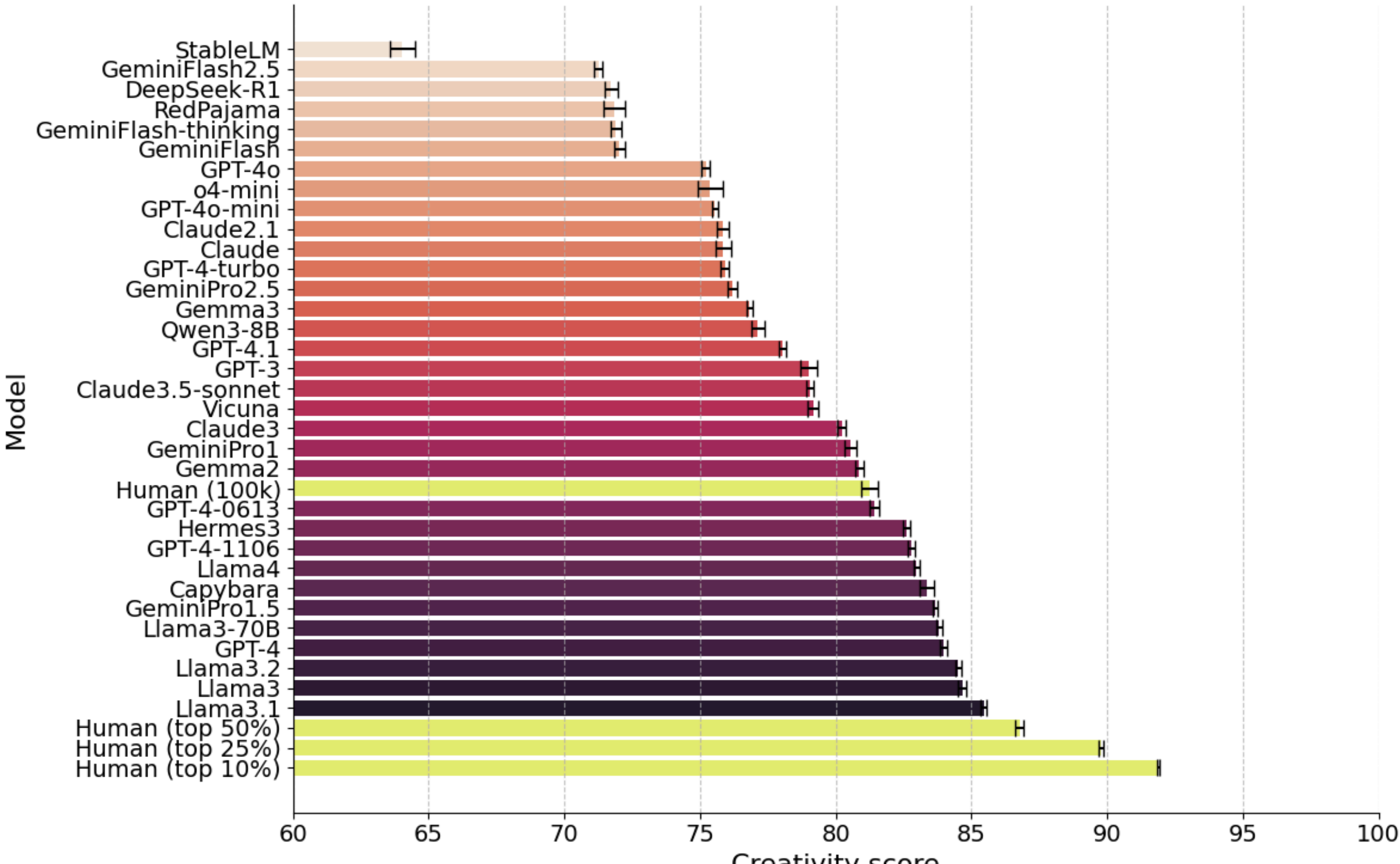

**Figure 2. Mean creativity scores for a wide range of large language models (LLMs) and human samples on the Divergent Association Task (DAT).** Models are ranked from lowest to highest mean score, with error bars indicating 95% confidence intervals. For humans, each bar represents the mean of a random subsample of 500 responses (n = 500), drawn either from the full distribution (N = 100,000) or restricted to the top 50% (N = 50,000), 25% (N = 25,000), or 10% (N = 10,000) of responses. For LLMs, each bar represents the mean of 500 model-generated responses.

To further contextualize these findings, **Figure 2** presents a comprehensive comparison of creativity scores across an expanded set of LLMs released between January 2023 and June 2025 alongside different segments of the human population taken from our sample.. Consistent with our main analyses, several leading LLMs now reliably exceed the average score of the general population. However, the most creative humans—those in the top decile, quartile and above median—still achieve higher DAT scores than any model of our curated list (see supplementary Figure S5, S6 and Table S1 for more details on statistical significance, response distributions and model specifications). This result underscores a persistent gap between artificial and human divergent thinking at the highest levels, despite rapid advancements in LLM design.

### Assessing the validity of the DAT across LLMs

To validate the models' compliance with the DAT instructions and to ensure their responses weren't arbitrary word distributions, we compared their performance to a control condition, which entailed prompting the LLMs to generate a list of 10 words, without specifying a need for maximal difference between the words. The findings, illustrated in Figure 3 reveal that, when prompted with DAT instructions, every model significantly outperformed the control condition. This result was taken as evidence for the adherence of the LLMs to the task of producing a maximally divergent set of words.

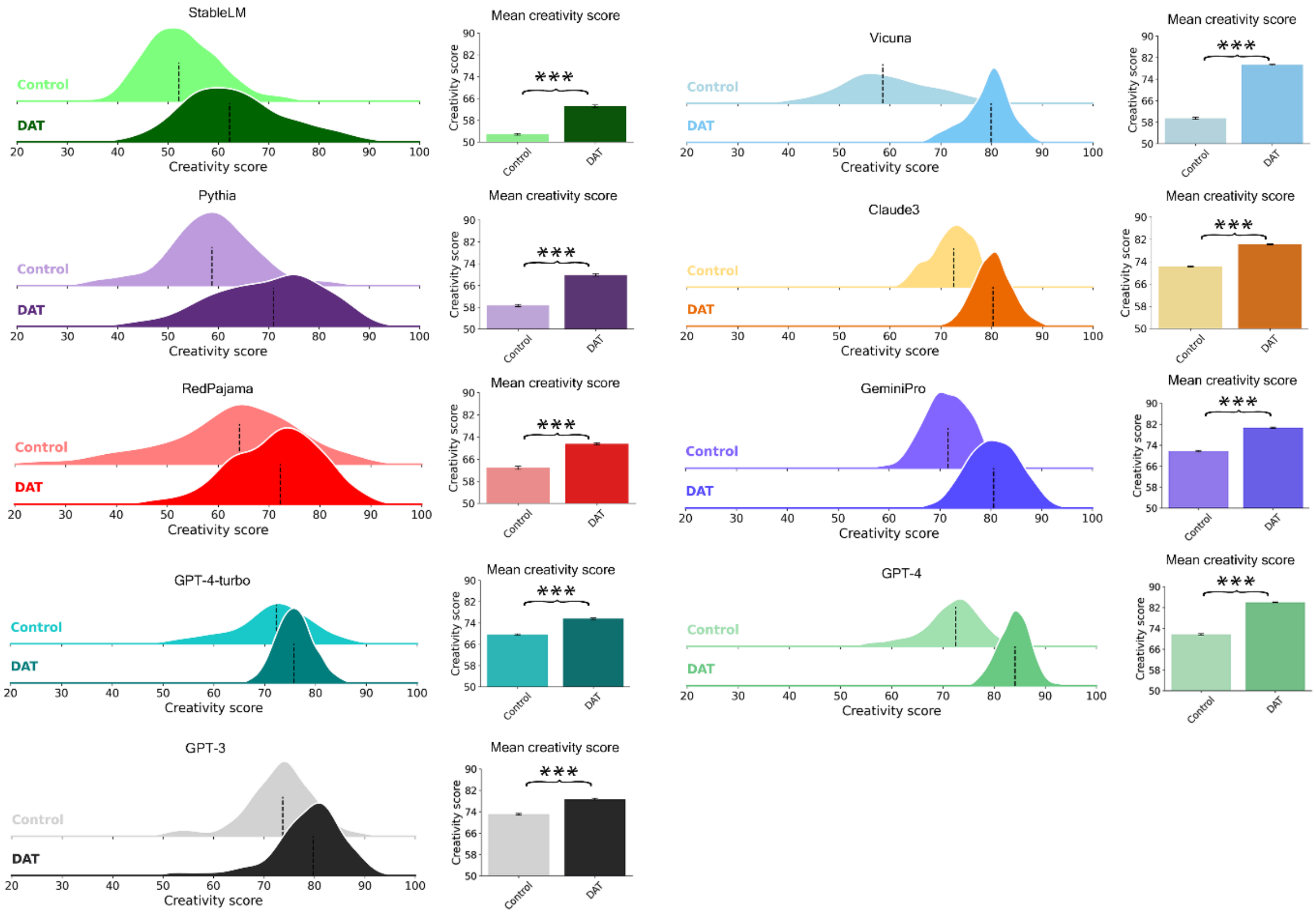

**Fig 3. DAT compared to the control condition across LLMs.** Performance of each model when being prompted with the original DAT instructions versus when being prompted to write a generic list of ten words. Each contrast is sorted in ascending order based on their mean performance in responding to the DAT instructions. ***: p<.001.

## The effect of model temperature on creativity scores

In order to evaluate the potential for modulating LLMs' creative performance via hyperparameter tuning, we explored the impact of adjusting the temperature value in GPT-4, the top-performing model. The underlying premise is that increased temperature would result in less deterministic responses, thereby yielding higher creativity scores. In line with this hypothesis, we observed a significant rise in DAT scores as a function of temperature (**Fig. 4A**), with a mean score of 85.6 achieved in the highest temperature condition (**Fig. 4B**). This mean score was higher than 72% of the human scores.

Notably, we found a reduced frequency of word repetitions as temperature increased, corroborating the notion that higher temperatures facilitate more diverse word sampling, whereas lower temperatures give rise to more deterministic responses (**Fig. 3C**). Interestingly, this pattern suggests that the superior performance of the top model is not simply attributable to the repetition of a well-optimized set of words (reflected in a high word count), but rather its ability to generate more and diverse responses.

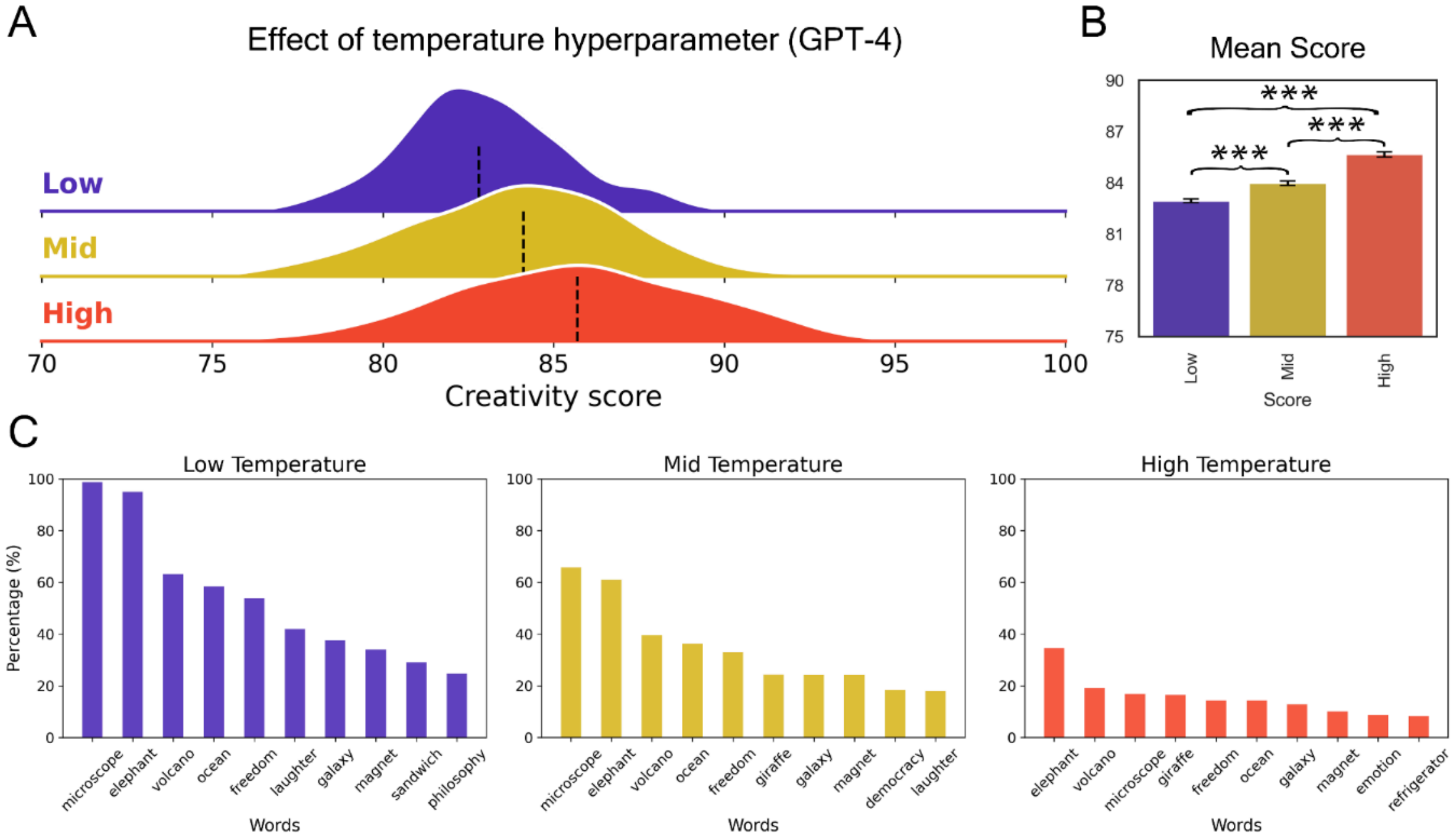

**Fig 4. GPT-4 creativity scores across temperature levels.** Varying performance across temperature levels in GPT-4 using the original DAT instructions. Each condition includes n = 500 generations. **(A)** Distributions of scores for each temperature level (Low: 0.5, Mid: 1.0, High: 1.5). Black vertical lines represent the median. **(B)** Barplot of the mean scores for each temperature level with results of the two-sided independent t-tests for each contrast. **(C)** Qualitative summary of the responses showing the 10 most frequent words across repetitions within each temperature condition. ***: $p<.001$.

## Exploring strategies to manipulate LLMs performances

We found that imposing specific strategies influenced LLM performance on the task, as illustrated by the performance-based ranking of strategies (**Fig. 5A-D**). To prompt the model to adopt different strategies in answering the DAT, we added a specification of the strategy to use at the end of the instructions, using the following sentence structure: *"[...] using a strategy that relies on meaning opposition | using a thesaurus | varying etymology*". All differences in means were statistically significant, with the exception of the contrast between the *Thesaurus* and *Basic Instructions,* highlighting the impact of strategy variations on LLM creativity scores. Interestingly, we observed that the Etymology strategy outperformed the original DAT prompt for both GPT-3.5 and GPT-4. This finding implies that these models exhibit higher DAT scores when explicitly prompted to use "a strategy that relies on varying etymology." Interestingly, although the strategy trends were similar across GPT-3.5 and GPT-4, we also noticed subtle differences between the two. Specifically, the Thesaurus strategy also outperformed the DAT in GPT-4.

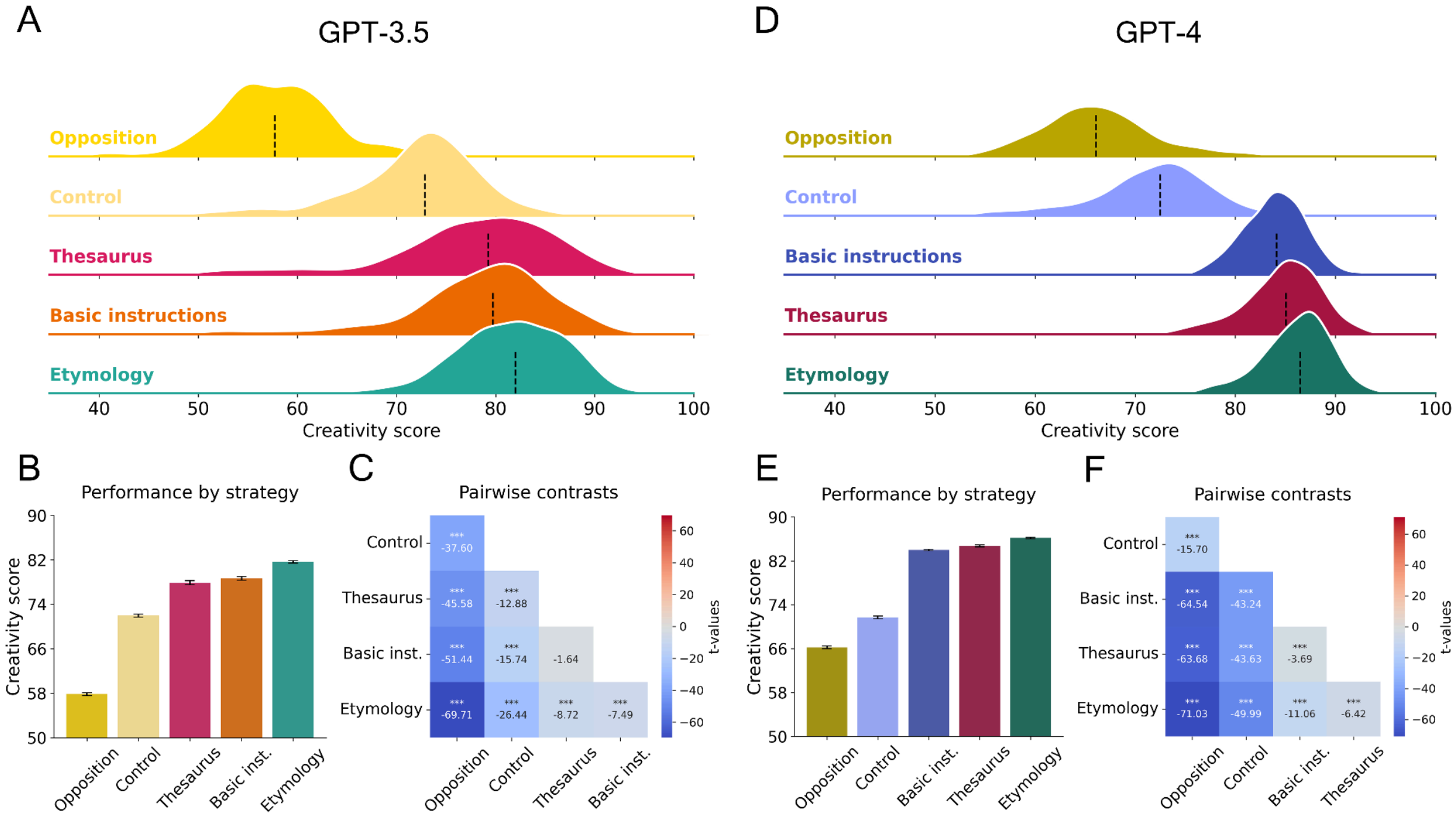

**Fig 5. Comparison of DAT scores for GPT-3.5 and GPT-4 across different linguistic strategies. (A, D)** Distribution for each strategy using a ridge plot of smoothed kernel density estimates for the two models. Black vertical lines represent the median. **(B, E)** Mean DAT score and 95% confidence intervals. **(C, F)** Heatmap of all contrasts, arranged in comparison to the highest performing strategy. *: p<.05, **: p<.01, ***: p<.001.

## Investigating LLMs' performance on creative writing tasks

Our exploration of LLM's ability to produce creative-like outputs extended beyond the DAT to a range of creative writing tasks designed to further interrogate the models' creative capabilities in relation to human generated corpuses. These tasks, including the generation of haikus (three-line poems), movie synopses, and flash fiction (brief narratives), were employed as complementary investigations to corroborate the DAT findings and provide broader evidence of the creative capacities of the examined LLMs. The three models that scored highest in the DAT (GPT-3.5, Vicuna, and GPT-4) were used to generate creative writing samples. In analyzing these creative outputs, we employed the Divergent Semantic Integration (DSI) to measure divergence across sentences, Lempel-Ziv Complexity for assessing unpredictability and diversity, and Principal Components Analysis (PCA) embeddings to understand thematic coherence and variance (see *Methods*).

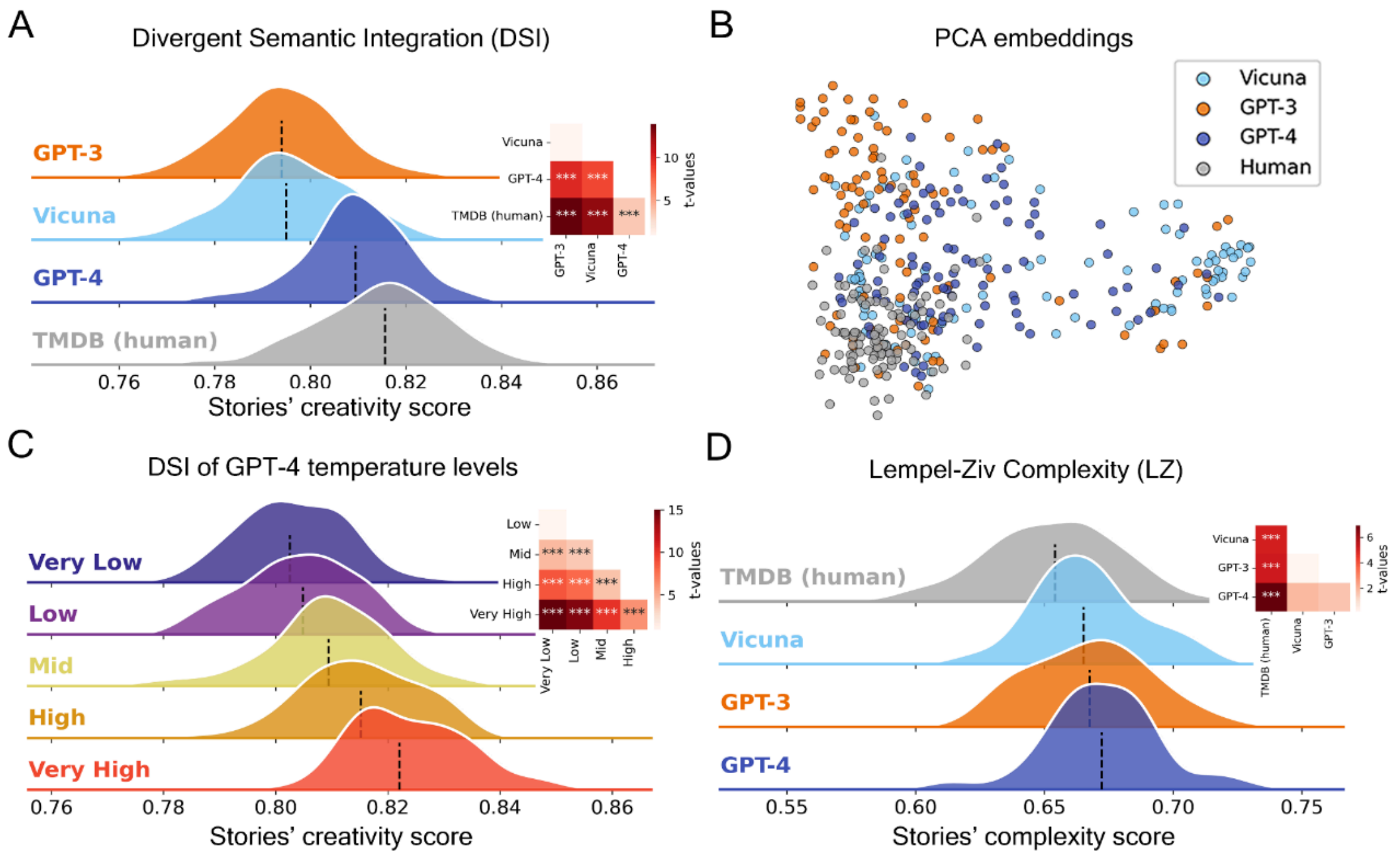

**Fig 6. Creative assessment of LLMs and human generated synopses.** Overview of the level of semantic divergence in synopses generated by humans and high-performing LLMs using different methodologies. **(A)** Distributions of DSI values across all models and human participants. **(B)** Scatterplot of the two-dimensional PCA performed on all synopses' embeddings. **(C)** Distributions of DSI values across temperature levels for GPT-4. **(D)** Distribution of normalized LZ complexity across models and human participants. *: p<.05, **: p<.01, ***: p<.001.

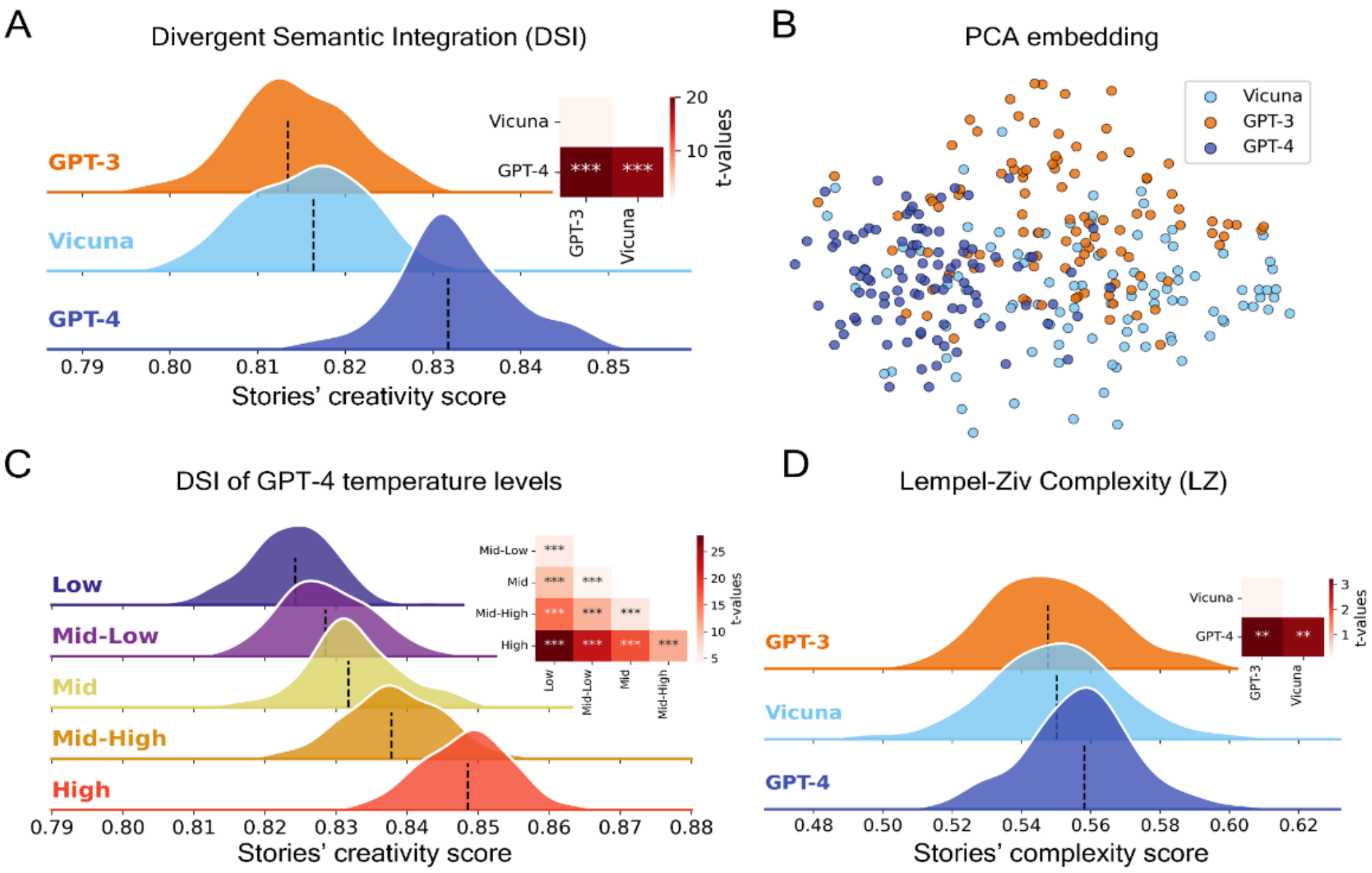

**Fig 7. Creative assessment of LLMs generated flash fiction.** Overview of the level of semantic divergence in flash fiction generated by high-performing LLMs using different methodologies. **(A)** Distributions of DSI values across all models. **(B)** Scatterplot of the two-dimensional PCA performed on all flash fiction embeddings. **(C)** Distributions of DSI values across temperature levels for GPT-4. **(D)** Distribution of normalized LZ complexity across models. *: p<.05, **: p<.01, ***: p<.001.

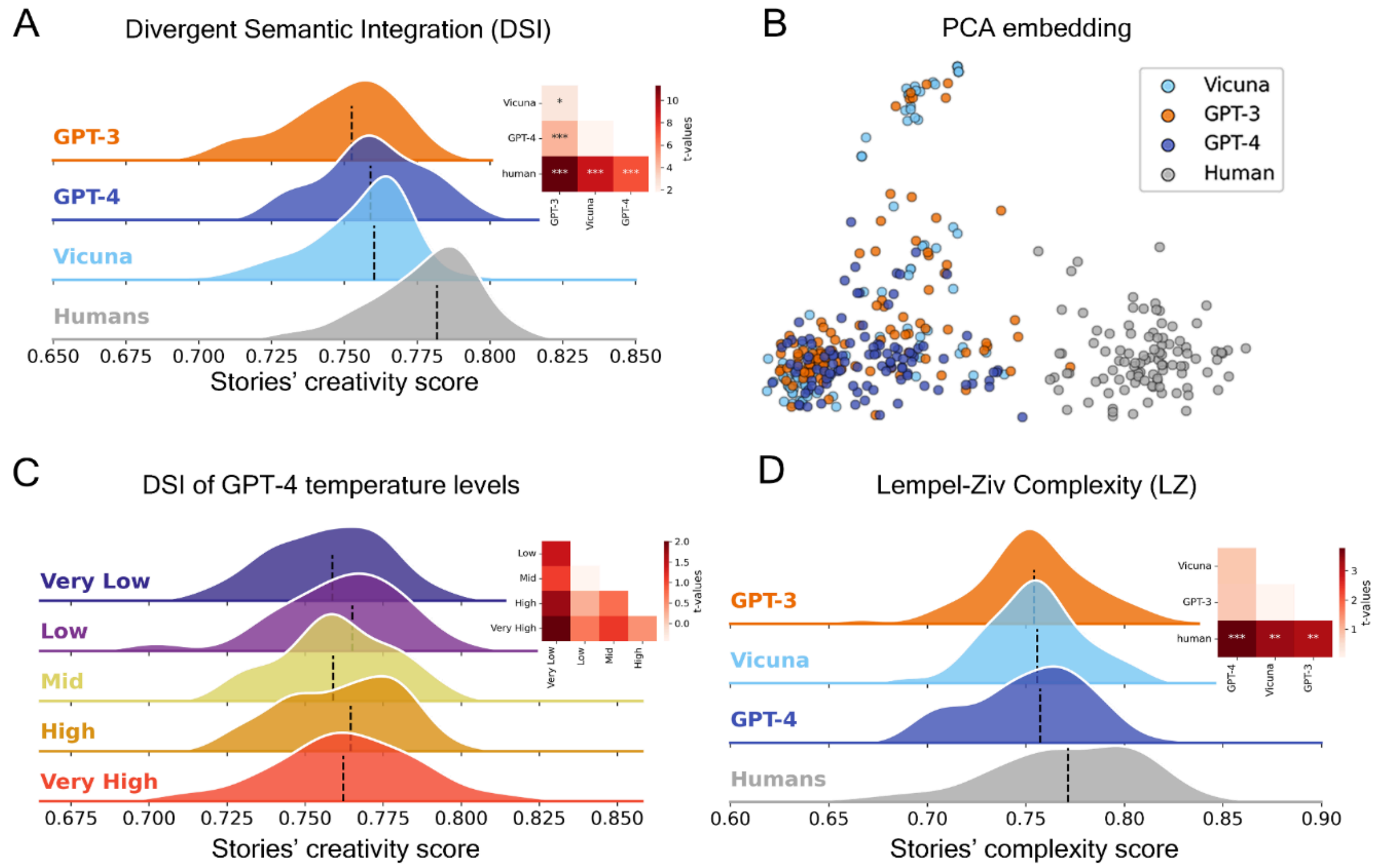

**Fig 8. Assessment of creativity on LLM and human generated haikus.** Overview of the level of semantic divergence in haikus generated by humans and high-performing LLMs using different methodologies. **(A)** Distributions of DSI values across all models and human participants. **(B)** Scatterplot of the two-dimensional PCA performed on all haikus embeddings. **(C)** Distributions of DSI values across temperature levels for GPT-4. **(D)** Distribution of normalized LZ complexity across models and human participants. *: p<.05, **: p<.01, ***: p<.001.

Our results indicate that GPT-4 consistently outperforms GPT-3.5 across all three categories of creative writing, as evaluated by Divergent Semantic Integration (DSI) (**Figs 6A, 7A, 8A**). Despite this, human-written samples maintain a significant edge in creativity over both language models. We also observe that the temperature parameter in GPT-4 heavily influences the DSI for synopses and flash fiction, with higher temperature settings correlating with increased creativity scores (**Figs 6C, 7C**). Interestingly, while temperature doesn't significantly affect the creative scores of haikus, it does play a more prominent role in longer writing formats, underscoring that such formats exhibit more pronounced differences in DSI scores in response to changes in temperature. While the overall DSI score variation across temperature settings in synopses appears modest, the differences are statistically significant ($p$ < .001) and become more pronounced in less structured formats like Flash Fiction (see Fig. 6). This suggests that task constraints modulate the impact of temperature on creative divergence.

A two-dimensional PCA embedding revealed distinct patterns, particularly when contrasting human responses to those of language models. In the case of both haikus and synopses, PCA reveals a clear separation between the embeddings of human-generated texts and those generated by LLMs. The clusters for different LLMs also occupy distinct regions in the embedding space. . Additionally, when PCA is applied to the flash fiction data, it effectively distinguishes the three different LLMs, as depicted in **Figures 6B**, **7B**, and **8B**.

In relation to Lempel-Ziv complexity scores, the pattern mirrors in most cases the performance order indicated by the DSI. Humans exhibit higher scores than LLMs for haikus, which is consistent with the DSI findings. However, humans' LZ scores are significantly lower than LLMs for synopses, in contrast to the DSI results.

This suggests the DAT is a useful tool for quantifying and comparing a specific ability for generating diverse sets of words across different LLMs and conditions. However, establishing the full psychometric properties and interpreting these scores in terms of 'creativity' or 'divergent thinking' analogous to humans requires further investigation. Investigating the underlying mechanisms and latent structures, which likely differ significantly between humans and LLMs even when producing similar outputs, is crucial for validating the DAT's broader implications in evaluating LLMs' potential to generate truly original text.

# Discussion

The aim of the present paper was to benchmark the performance of a wide range of LLMs on a straightforward and validated creativity test, while comparing their scores to a large cohort of human responses ($N$ = 100,000). Additionally, we aimed to modulate the creative performance of the highest-scoring models by adjusting the temperature level and the strategic approach employed by the LLMs in response to the DAT instruction. State-of-the-art LLMs exhibited remarkable proximity to human performance levels in the creativity assessment; the DAT scores of GeminiPro were statistically close to human performance, while GPT-4 exceeded it. It's crucial to understand that this finding is nontrivial as LLMs do not directly access all semantic distances between word pairs; instead, they depend on iterative transformations of latent representations, which differ from those used in the DAT computations.

Our results illustrate how targeted prompt design allows for the manipulation of LLMs' creative outputs, as assessed by the DAT. To strengthen our findings, we also demonstrated that performance on the DAT aligns with creative scores across multiple writing formats, as measured through DSI and LZ. This suggests that the chosen metrics have potential for broad applications in assessing other types of creative outputs, either through matrix operations (cosine similarity) for assessing semantic distance or compression algorithms for assessing redundancy.

## LLMs Surpass the Average—But Not the Most Creative Humans

One highlight of the present study certainly lies in the finding that, on average, GPT-4 performs reliably better than 100,000 humans on the divergent association task, although its highest scores still fall short of the best human performers. This finding supports the claim made by OpenAI that GPT-4 is more creative than its predecessor, but it also challenges the assumption that language-based tasks are sufficient to understand human creativity in general. Moreover, the performance of GPT-4-turbo, which significantly decreases compared to its predecessor GPT-4, indicates that efficiency improvements or cost reductions might come at the expense of increased redundancy across the model's responses, suggesting a trade-off between diversity and resource optimization in the development of language models. Recent investigations have contrasted human and artificial creativity employing the Alternative Uses Task (AUT), revealing for instance that humans surpass GPT-3.5 in creative output (*22*). In contrast, another study using the same task but with a different scoring approach found that both GPT-3.5 and GPT-4 outperform humans on average (*39*). A separate study evaluating multiple models found that their scores on the AUT are similar to human performance, with some evidence that GPT-4 can exceed human originality (*24*). A classical battery of creativity tests, the Torrance Tests of Creative Thinking, was also used to benchmark GPT-4 performance and found that it scored within the top 1% for originality and fluency (*38*). One study also assessed the DAT in GPT-3.5 and GPT-4 compared to a human sample, showing that both models outperform humans on average (*37*). Our findings expand upon these insights by (i) juxtaposing human responses with a more extensive array of LLMs, (ii) exploring multiple creativity-related metrics which show potential for comparing LLMs and humans (DAT, DSI, and LZ complexity), (iii) comparing DAT benchmarking with performance on several creative writing tasks, providing convergent evidence for its validity as a proxy for creative writing evaluation in LLMs, (iv) using an unprecedented large human dataset ($n$ = 100,000), all English speakers and balanced for age and sex, (v) verifying for adherence to the DAT instructions through comparison with a control condition, (vi) exploring the effect of hyperparameter tuning (temperature) and prompt design strategies, and (vii) sharing code that both uses direct calls to the API of all closed source models, as well as scripts to run open-source LLMs locally. Despite widespread concern that AI could imminently replace creative professionals (like writers, for instance), our results suggest that such fears remain premature. The persistent gap between the best-performing humans and even the most advanced LLMs indicates that the most demanding creative roles in industry are unlikely to be supplanted by current artificial intelligence systems.

**LLM creativity can be manipulated through prompt design and hyperparameter settings**

Our comparison of the DAT versus control conditions reaffirms this observation, with all tested LLMs demonstrating a significant increase in DAT scores when instructed explicitly to generate a list of maximally different words compared to merely listing random words. This distinction underscores the sensitivity of LLMs to the nuances of task instructions and their capability to adjust their output based on these specifications. Moreover, the performance of LLMs varied markedly when exposed to different strategies. As expected, when prompted to use the opposition strategy, the models' performance significantly decreased, as opposing words (e.g. "light" and "darkness") have a relatively low semantic distance. We also found that when explicitly prompted to use words with varying etymology, both GPT-3.5 and GPT-4 outperformed the original DAT prompts, suggesting the potential for enhancing semantic divergence by referring to the roots of words. These observations align with recent findings showing significant increases in GPT-3.5 performance on the AUT (Alternative Uses Test) when prompted to adopt a two-phase approach of brainstorming followed by selection, surpassing human creativity scores in some instances (*49*). Thus, our results, in concert with these findings, indicate that manipulating prompts can be a powerful tool for modulating the creative performance of LLMs. The efficacy of specifying strategies raises intriguing questions about potential parallels in human creative processes. It is plausible that humans, while responding to the DAT, implicitly or explicitly employ a mix of strategies to generate their responses. Future research would benefit from exploring this dimension, systematically comparing human strategic approaches with those we can program into LLMs. For example, studies could verify whether changing the instructions given to humans or LLMs result in similar changes in performance. Such comparative analyses could further our understanding of how strategy manipulation can be leveraged to enhance the creative performance of both LLMs and humans.

In addition to prompting strategies, hyperparameter tuning was found to significantly bolster the performance of LLMs, particularly GPT-4. An increase in temperature led to a substantial rise in DAT scores, with the highest temperature condition surpassing the mean creativity score of a significant portion of human participants. This increase in semantic divergence aligns with the concurrent decrease in word repetition frequency, suggesting that higher temperatures indeed diversify word selection, steering LLMs away from deterministic responses. Although low-temperature settings result in frequent repetition of certain high-probability words (e.g., "microscope" and "elephant"), this reflects the model's inherent token selection mechanism rather than a lack of semantic diversity. Unlike humans, LLMs operate on probability distributions that favor the same high-probability tokens under deterministic conditions, given their lack of memory between repetitions of the task. However, as temperature increases, the determinism of the response decreases, allowing the model to explore a broader range of potential continuations, which can lead to more diverse outputs. Nonetheless, the overall semantic relationships—captured by the DSI—remain informative of creative performance. This modulation of creativity via temperature adjustment presents an interesting parallel to the divergent (variation) phase of human creativity (*19*, *20*). Increasing the temperature broadens the solution space from which the LLM can draw, mirroring the expansive ideation characteristic of the human divergent process. However, the potential proficiency of LLMs in the ensuing convergent phase of the creative process, where the selection of the most useful and original ideas occur, remains under-explored. As we advance our understanding of LLMs and their creative capabilities, this represents a key area for further inquiry into associative thinking (*53*).

Additionally, our findings delineate a discernible divergence between prompt adherence, as demonstrated by the proportion of correctly formatted responses (Fig. 1c), and creativity enhancement. Creativity extends beyond the mere capacity of the model to accomplish the instruction in the stipulated format.

**Insights from the comparable semantic creativity of humans and machines**

The parallel in outcomes between LLMs and humans on the DAT task, despite the obvious differences between the underlying processes by which each completes the task, calls for inquiry into whether state-of-the-art tools for divergent thinking assessment serve as trustworthy markers of creativity. This concern is compounded by the fact that many of these tools were developed under assumptions specific to human cognition—like constraints on memory and contextual generalization—which may not translate to LLMs, potentially undermining the validity of such assessments in this new context (*54*)

Incidentally, this question has been the focus of a few recent studies. One study delves into the notion of embodied cognition and symbolic processing in LLMs, arguing that these notions may be more intertwined than previously assumed. It suggests that LLMs do not emulate the cognitive processes of humans, lacking similar embodied experiences that ground language processing (*55*). This nuanced understanding is further reinforced by another study, revealing specific areas where ChatGPT's language usage patterns diverge from humans', such as disfavoring shorter words for less informative content and not using context for syntactic ambiguity resolution (*56*). A third study notes the dissimilarities between LLMs and humans in learning hierarchical structures and abstractions, essential facets of linguistic processing (*57*).

Furthermore, as LLMs adhere to the nuances of prompt deployment, the exploration of more refined models of their "internal processes", such as quantifying attention flows (*58*), could offer a more nuanced understanding of their word production, and consequently, their creativity. Critically, in the context of language understanding, assessing the competence of LLMs requires more than simply evaluating their output performance (*59*). Instead, understanding the "internal processes" of LLMs provides a more accurate gauge of their potential to model human language, highlighting possible differences in cognitive processing mechanisms in brains and machines. In fact, further quantifying the alignment of biological and artificial neural networks is a promising area of research, as it has been shown that both can be analyzed using vector embeddings (*60*), with empirical paradigms looking at similarities between language models and language areas of the brain (*61*).

**From divergent association to creative writing**

Upon interpreting the DAT performance, we transposed our methodology to longer-form text in order to capture whether the creative ability of LLMs would remain when creating short narratives or poetry. We employed Divergent Semantic Integration (DSI) and Lempel-Ziv (LZ) complexity scores to assess the creativity of various LLM-generated written output, encompassing haikus, synopses, and flash fiction. Our findings affirmed a parallel between high performers on the DAT and those exhibiting superior creativity in long-form writings, thus validating our metrics and demonstrating the interconnection between these distinct facets of creative expression. Moreover, increasing the temperature of GPT-4 led to a marked enhancement in creativity scores, most prominently in synopsis and flash fiction tasks, affirming the replicability of hyperparameter tuning effects from the DAT onto creative writing. Intriguingly, this temperature effect appears more potent for longer text formats, such as flash fiction, compared to shorter ones, such as haikus. Perhaps temperature does not affect the inference process when producing short texts, or the DSI does not capture the effect of temperature in such a format. It remains possible that temperature increases the creativity of single-sentence outputs, but that this effect is not detected by our current method. In comparing humans and LLMs in creative writing, the superior performance of humans in haikus and synopses may be explained by the fact that the human writers likely had advanced creative writing skills, unlike the general population used for the human DAT scores. While our results show that LLMs can approximate human creativity on certain automated metrics, LLM-generated stories tend to pass Torrance Test of Creative Writing (TTCW) at significantly lower frequencies—3 to 10 times less—than those produced by professional writers (*62*). This reinforces the view that current automated measures must be complemented by human-based evaluations to fully capture the nuances of creative writing.

To explore broader patterns in writing style and thematic variance across models, we applied PCA to text embeddings. The resulting visualizations revealed that texts generated by different LLMs occupy distinct regions in the embedding space, suggesting variation in stylistic or semantic patterns. In the case of synopses and haikus, human-generated texts also formed a separate cluster from LLM outputs. While PCA does not serve as a metric of creativity, it offers a complementary perspective by visualizing structural differences in language use. We present these results not as a tool for assessing creative quality, but as an exploratory method for illustrating diversity in textual production across models and human authors.

An intriguing observation arose in relation to haikus. Since a haiku traditionally centers on imagery from nature, the higher LZ complexity and DSI scores observed in human-generated haikus may suggest that humans adhered less strictly to the rules compared to LLMs. We tested this by plotting the cosine similarity between word embeddings across all haikus and the word "nature" (see Fig. S1). The results indeed show that humans and LLMs at a high temperature appear to diverge from the canonical requirement for nature imagery, which could explain their higher LZ complexity and DSI scores.

**Divergence and complexity metrics as promising tools for LLM creativity benchmarking**

Our findings underscore the significant potential of divergence and complexity metrics as novel tools for assessing and benchmarking the creativity of LLMs. To the best of our knowledge, the present study is the first to combine several distinct word embedding models (BERT, GLoVe, and OpenAI's embeddings API) to assess and benchmark semantic creativity of LLMs. By combining the DAT, DSI and LZ complexity scores, the present study reveals nuanced and complementary insights into the creative capacities of LLMs across a variety of tasks (DAT, haikus, synopses, flash fiction). While the DAT and DSI focus on the semantic distance between word or sentence-level embeddings, LZ complexity captures the semantic richness by measuring the diversity among a set of words. Therefore, the distinct patterns of LLM performance based on these metrics, observed across different creative writing formats and temperature parameters, validate their utility as efficient LLM benchmarking tools. While our comparisons demonstrate that LLMs approach human performance on certain divergence metrics, our results must be viewed in the context of a broader literature on homogenization in model outputs (*63–65*). These studies suggest that improvements in language generation quality and alignment may come at the cost of reduced diversity, highlighting a trade-off that warrants further investigation.

Current benchmarks focus on responses to closed-ended scenarios, such as finding the correct answer to sets of multiple-choice questions that constitute exams. Answering these requires convergent thinking, in which multiple inputs are assessed before choosing the most appropriate output. Convergent thinking tasks are easy to score, and easy to use as benchmarks, because there is a single best response. Complementing these convergent thinking benchmarks with divergent thinking tasks and other measures of creativity, as shown here, may give a more holistic view of LLM performance. Indeed, given that current models are prone to hallucination, they may be particularly suited for benchmarks based on divergent thinking. Hallucination in divergent thinking tasks can still result in a good (but different) response, whereas hallucination in convergent thinking can lead to an objectively wrong answer.

## From competition to collaboration

Moving beyond a simple comparison of creative performance between humans and LLMs, our findings open intriguing questions. Could we use LLMs to build individualized models of creative thinking? Can these models enable a deeper understanding of the human creative process? The potential of LLMs to mirror, and even enhance, human creativity opens up exciting possibilities for human-machine collaborations in creative endeavors (*44*, *45*, *47*, *49*, *66*). In a related empirical study, LLMs were shown to effectively support emerging writers, particularly in translation and revision stages of the writing process, though challenges remain in fostering ideation and originality (*67*). More broadly, a recent meta-analysis of 106 studies revealed that while human–AI collaboration often underperforms the best of either alone, it shows notable advantages in content creation tasks, highlighting the context-dependent nature of such synergies (*68*). Exploring the potential intersections between human and machine creativity, it becomes essential to think about how this convergence can offer holistic insights into creativity as an experiential and computational phenomenon. Future research building upon the framework we propose here may shed light on the need for better synergy between the phenomenology of creativity and its implementation in generative models (*69*). Indeed, the development of generative modeling techniques can be pursued as computational models of lived experiences, with the aim of establishing 'generative passages' between first-person accounts and their third-person descriptions (e.g. neural processes). This would allow us to gain more explicit formalizations of creativity (i.e. as lived experiences, not only linguistic artifacts). Thus, our study not only broadens the horizon of LLM evaluation but also envisions a future where human and machine creativity coalesce, through practices like computational phenomenology, to drive innovation responsibly.

## Limitations and perspectives

While our research provides valuable insights into the creativity of LLMs, several limitations and caveats are worth noting. Firstly, properties such as architecture and size were not publicly available for some of the closed source models we used. This restriction hampered our ability to draw definitive conclusions about the contribution of specific features and configurations to the observed performance. Secondly, it is noteworthy that the fast-paced development in the field of LLMs would require continuous updates of the analyses presented here. To this end, we have made the associated code and tools available to the broader AI and creativity research communities, facilitating ongoing assessment of new and updated models. By incorporating these tools into the standard toolkit for LLM evaluation, we can promote a more holistic, nuanced understanding of LLM performance, thus driving advancements in model development and refinement. Thirdly, leveraging semantic distance as a metric inherently constrains the evaluated scope of creativity in texts like poetry, synopses, or fiction. It is plausible that a text can manifest novel ideas using semantically close words. Nevertheless, prior studies validate a notable correlation between human creativity ratings and DSI scores (*35*), reinforcing the notion that DSI effectively captures components of semantic creativity that align with human judgment. However, in order to demonstrate the unambiguous alignment between human and language models in terms of semantic creativity, a comprehensive benchmark using both automated scoring procedures and expert judgments would probably be ideal. In this sense, exploring other aspects of creativity in future work, such as convergent thinking, and considering constraints such as usefulness and novelty, will paint a more comprehensive picture of LLMs' creative abilities. Fourthly, considering the subjective nature of creativity, future research must rigorously incorporate human evaluations, particularly when assessing LLM outputs. While automated metrics like DSI have shown correlation with human ratings for human-generated text, their direct applicability and validity for LLM-generated content require specific investigation, as LLMs may exhibit unique stylistic or structural properties. A preliminary comparison between human expert ratings and automated ratings by GPT-4 on the same creative texts revealed that GPT-4 raters exhibited higher internal consistency than human raters, while the agreement between GPT-4 and human ratings was lower than that observed within each group. Nonetheless, we found a positive correlation between human and GPT-4 ratings overall. Moreover, using GPT-4 ratings as a proxy, we demonstrate that the ratings assigned to LLM-generated flash fiction stories allow us to rank different models in the same order as the Divergent Semantic Integration (DSI) metric, with GPT-4 achieving the highest ratings (see **Fig S4**). These findings indicate that automated GPT-4 ratings can capture model performance trends similar to human evaluations, although they should be interpreted with caution due to the inherent differences in rating consistency between machines and human experts. Lastly, the exact knowledge cutoff dates for many commercial LLMs remain unclear due to both proprietary restrictions and the dynamic nature of aggregated, continuously updated datasets—as exemplified by models like RedPajama and StableLM. The lack of transparency regarding the exact training data of most commercial models represents an uncontrollable confound—particularly in determining whether models were exposed to specific prompts such as the DAT. We explicitly acknowledge this limitation in the context of assessing prompt familiarity and prior exposure.

By employing complementary metrics of creativity which rely on distinct embedding methods, we provide a thorough assessment of semantic creativity in both LLMs and 100,000 humans across various language production tasks. We observed that the top performing models reached and, in some cases, even surpassed human scores on the DAT. Furthermore, we found that DAT creativity scores were modulated by prompt design and model temperature. Importantly, the observations obtained with a simple semantic creativity test were found to be generalizable to richer and more complex creative writing tasks, including poetry, movie synopsis, and short fiction.

While LLMs are often benchmarked using a wide variety of tests typically used to assess human performance, creativity — a cornerstone of human cognition — remains widely unexplored in machines. Beyond the specifics of the observations reported here, our methodological framework sets the stage for creativity metrics to become one of the standard measures in assessing the performance of future models. While our focus was on LLM creativity, the questions raised here extend to all forms of generative AI, whether generating images, videos, or music.

This research also reframes our understanding of divergent creativity by encouraging more granular inquiries into the distinctive elements that constitute human inventive thought processes, compared to those that are artificially generated. More generally, our methodology offers a promising foundation for future research at the intersection of computational linguistics and creativity.

## Materials and Methods

### Experimental Design

*Standard DAT protocol in humans*

To evaluate creativity in humans, we employed the Divergent Association Task (DAT; (*29*)), which involves participants generating 10 words that are maximally different from one another in meaning and usage. The difference between the words is computed using semantic distance, as determined by the cosine similarity between embedding vectors from the GLoVe model, which convert words into numerical vectors in a high-dimensional space (*33*). The first 7 valid (i.e., properly spelled) words are used, which allows occasional misspellings to still result in a valid score. The average of the pairwise semantic distances across the 7 words is then used to derive the DAT score (*29*). Scores typically range between 50 and 100, with higher scores indicating more semantic distance and higher creativity. Although the scores can theoretically range between 0 and 200, the practical limits for the corpus used range between approximately 6 and 110.

*Adaptation of the DAT to function as a chat prompt*

To use the DAT with LLMs, we adapted the original DAT instruction to function as a chat prompt, cueing the LLMs to output a structured response allowing us to quantify DAT scores in an automated fashion. We used the following prompt:

```
      Please enter 10 words that are as different from each other as possible, in all meanings and uses of the words. Rules: Only single words in English. Only nouns (e.g., things, objects, concepts). No proper nouns (e.g., no specific people or places). No specialized vocabulary (e.g., no technical terms). Think of the words on your own (e.g., do not just look at objects in your surroundings).  Make a list of these 10 words, a single word in each entry of the list.
```

*Control prompt*

To ensure that the LLMs generated responses based on the DAT instructions rather than random distributions of words, we incorporated a control condition in which we asked the models to simply output 10 words without further instructions (prompt: "make a list of 10 words"). This approach allowed us to verify LLM adherence to the task guidelines and gauge the validity of their creative outputs.

*DAT with strategies*

Furthermore, to investigate whether imposing a particular strategy influences LLM performance in the task, we introduced variations to the DAT instructions in which we probed three specific strategies: etymology (focusing on the root of the words), thesaurus (concerned with synonyms), and meaning opposition (listing words with opposite meaning). This manipulation was aimed at providing insights into the adaptability and flexibility of LLMs in creative problem solving, and at assessing the impact of prompting on the performance in the task.

*Prompting creative writing*

To investigate the relationship between performance on the DAT and creative abilities, we assigned LLMs a variety of creative writing tasks, encompassing the creation of haikus, synopses, and flash (short) fiction. Haikus consist of seventeen syllables distributed over three lines in a five-seven-five pattern and typically encapsulate vivid imagery of nature. Synopses involve summarizing a film's plot succinctly, while flash fiction represents a literary genre characterized by extremely concise storytelling. Movie synopsis is of course not a direct reflection of the creativity of the movie itself. Yet, writing a good movie synopsis demands specific creative writing skills—such as narrative framing and stylistic originality—aimed at crafting a brief yet captivating description designed to spark curiosity and entice viewers to watch the film. We posed challenges to the LLMs to "Invent a haiku", "Invent the synopsis of a movie", or "Invent a flash fiction", with a strict word limit of 50 words for synopses and 200 words for flash fiction stories. Haikus, due to their inherent syllable restriction, required no additional constraints. The length of the generated texts was later verified (see Assessment of the structure of creative writing section).

*Manipulating LLM temperature*

Temperature is a hyperparameter of LLMs that refers to the degree of randomness in the word sampling process and can be regarded as a reflection of the exploration/exploitation tradeoff in creativity. A higher temperature in LLMs can be seen as fostering exploration, allowing for more creative but potentially less coherent outputs, while a lower temperature leans towards exploitation, generating text that is more predictable and contextually accurate. By adjusting the temperature across three levels, we can control the degree of randomness in the model's word sampling method, thus allowing for either more or less constrained text generation. This results in more deterministic responses at low temperatures and less deterministic responses at high temperatures, serving as a proxy for evaluating variability in creative behavior and the responsiveness of LLMs to parameter adjustments. Outside of the analysis specific to temperature, all other results were collected using the default temperature values for each model (see Table 1).

## Human participants and LLMs

*Demographics of human participants*

A total of 100,000 human participants (50% men, 50% women) were randomly selected from a larger study (*70*), with 20% from each age group (18 to 29, 30 to 39, 40 to 49, 50 to 59, and 60 and over). They were informed about the study from news articles, social media, or word of mouth, and came from the United Status (n = 79,832) and other English-speaking countries: the United Kingdom (n = 8,131), Canada (n = 7,601), Australia (n = 3,808), and New Zealand (n = 628). All participants were recruited directly via the official Divergent Association Task (DAT) website and received the same standardized writing prompt as defined in the original DAT study, ensuring consistency in prompt exposure and response evaluation.

*Selected LLMs*

The training and fine-tuning procedures (as disclosed publicly) of the LLMs used in this study are summarized in **Table 1**. Our selection encompasses popular AI products such as GPT-3.5, GPT-4 and GPT-4-turbo by OpenAI, Claude3 by Anthropic, and GeminiPro by Google, but also covers lesser-known open-source models such as Pythia by EleutherAI, StableLM by StabilityAI, RedPajama by Together, and Vicuna by NousResearch. We systematically used versions of each model fine-tuned on instructions, i.e. models that have been tweaked for better compliance in a conversation setting. These chat-enabled LLMs perform more or less correctly at the kind of "zero-shot learning" task we conducted. The selected models vary, among other things, in size, number of training tokens, fine-tuning methodology, temperature settings, and licensing conditions. The selection was not systematic per se, but it was intended to foster a wide comparison of performance across these different characteristics. We used default values for the top-p parameter, which controls the cumulative probability distribution from which the model selects its next token.

| Model name | Organization | Model ID | Model size | Fine-tuning | Temp range/ default | License | Source | Year of release |
|---|---|---|---|---|---|---|---|---|
| GPT3 | OpenAI | gtp-3.5-turbo | unknown | RLHF | 0-2 / 1 | Paid access | OpenAI API | March 15, 2022 |
| GPT4 | OpenAI | gpt-4-0314 | unknown | RLHF | 0-2 / 1 | Paid access | OpenAI API | March 14, 2023 |
| GPT4-turbo | OpenAI | gpt-4-0125-preview | unknown | RLHF | 0-2 / 1 | Paid access | OpenAI API | January 25, 2024 |
| Claude | Anthropic | claude-3-opus-20240229 | unknown | RLAIF | 0-1 / 1 | Paid access | Claude API | February 29, 2024 |
| Pythia | EleutherAI | oasst-sft-4-pythia-12b-epoch-3.5 | 12B | RLHF | 0-1 / 0.7 | Apache 2.0 | Open-Assistant huggingface | April 3, 2023 |
| GeminiPro | Google | n.a. | unknown | unknown | n.a. | Free access | Google Cloud Platform API | December 6, 2023 |
| StableLM | Stability | stablelm-tuned-alpha-7b | 7B | RLHF | 0-1 / 0.7 | CC BY-SA-4.0 | StabilityAI huggingface | April 20, 2023 |
| RedPajama | Together-Computer | RedPajama-INCITE-Chat-7B-v0.1 | 3B | RLHF | 0-1 / 0.7 | Apache 2.0 | Together-Computer huggingface | May 5, 2023 |
| gpt4-x-vicuna | Nous-Research | gpt4-x-vicuna-13b-ggml-q8_0 | 13B | RLAIF | 0-$\infty$ / 0.8 | GPL | TheBloke huggingface | May 20, 2023 |

**Table 1. Selection of Large-Language Models (LLMs) and their technical specifications.**

## Data Collection

For data collection with GPT-3.5 (*71*), GPT-4 (*17*), GPT-4-turbo (*17*), Pythia (*72*), and Claude3 (*73*, *74*), we made calls to the official APIs (see Table 1). We used the Google Cloud Platform Python Software Development Kit to run GeminiPro (*75*). Since model weights are publicly available for StableLM (*76*), RedPajama (*77*) and Vicuna (*78*), we collected responses running inference on the Digital Research Alliance of Canada compute cluster (alliancecan.ca/en), using NVIDIA V100 Volta (32 GB) GPUs.

Because each inference instance – or "chat session" – depends on a distinct random seed, multiple iterations of the same prompt using a different "chat session" lead to different responses. Therefore, we collected 500 samples for each prompt starting a new conversation with every iteration across all DAT conditions (DAT control, strategy, and temperature). For the creative writing tasks, we gathered a set of 100 examples for each creative writing prompt (haiku, synopsis, and flash fiction) from the LLMs using the same protocol.

To compare the performance of LLMs against human creative writing, we extracted texts from two established online resources. Synopses, succinctly encapsulating movie plots, were sourced from The Movie DataBase (TMDB), an accessible platform with a convenient API. To ensure generalizability, we randomly sampled synopses from a large corpus rather than selecting specifically highly creative writing. Our goal was to evaluate typical synopsis writing with respect to divergent semantic integration, not to benchmark peak creativity. For the haiku task, human-generated examples were obtained from Temps Libre, a dedicated online platform that serves as a repository for this traditional form of poetry. This diverse collection of human creative output served as a benchmark for comparing and evaluating the creative abilities of LLMs.

## Statistical Analysis

### *Scoring of the Divergent Association Test*

To ensure a robust estimate of the models' creative performance, we computed the mean DAT scores from 500 repetitions of the task after excluding answers with less than 10 words or which were otherwise incomplete. The primary reason for rejecting samples was mainly the model's inherent incapacity to generate a response that complies to the given instructions. We calculated the ratio of valid repetitions on the total number of repetitions to assess each model's compliance to instructions (prompt adherence). We also assessed the number of occurrences of every unique word (word count) across repetitions to capture the answers that are most representative of the models' performance.

### *Assessment of the structure of creative writing*

To ensure an equitable comparison between the creative writing samples produced by LLMs and humans, we first confirmed the compliance of these samples with the required low-level features. For the haiku task, we verified the adherence of each entry to the traditional 5-7-5 syllable structure, a cornerstone of this poetic form. Similarly, in the synopsis and flash fiction tasks, we ensured parity in the linguistic output by adjusting for the number of words and matching the mean and standard deviation of all distributions under scrutiny. This step allowed us to mitigate the confounding influences of variance in word count and focus on the quality and divergent integration of ideas within each submission.

### *Divergent Semantic Integration*

To compute Divergent Semantic Integration (DSI), we used Bidirectional Encoder Representation from Transformer (BERT) due to its ability to generate context-dependent word embeddings, as has been recommended (*35*)). First, the text was stripped of stop words and punctuation then was tokenized into individual words or morphemes. The text was further divided into sentences, from which the DSI score was calculated by determining the cosine similarity between every pair of successive word embeddings. These distances were averaged, effectively measuring the integration of diverse ideas within the text. In the calculation of the DSI scores, we specifically selected and combined layers 6 and 7 of the neural network to extract word embeddings that reflect relevant semantic and syntactic information (*35*). Research validated the use of these layers due to their high correlation with human creativity ratings.

### *Lempel-Ziv complexity of creative stories*

We used the Lempel-Ziv complexity algorithm (*79*) to examine the complexity of a text as an additional characteristic of creative writing outputs. This measure was initially developed for the purpose of lossless data compression; the modified Lempel-Ziv complexity evaluates the compressibility of a signal, which, in this instance, is a collection of text strings (rendered from a series of bytes). The compression algorithm operates on the principle of detecting repeated substrings from left to right. When a repetition is identified, the text can be replaced with a reference to its earlier occurrence, which results in a reduction of the text's size. The complexity score is defined as the number of unique substrings, which we normalized using the length of the text (*80*). We suggest that this method is capable of tracking divergence since a wide range of semantic content would result in fewer repetitions in the text, which effectively gauges the diversity and richness of the text.

Our rationale for including LZ complexity was twofold. First, it provides a complementary perspective to DSI by relying solely on the raw text structure, independent of any deep learning–based embeddings. Second, we hypothesized that reduced redundancy (i.e., higher complexity/lower compressibility) could serve as a proxy for lexical or structural diversity, potentially reflecting the novelty aspect of creativity.

We fully acknowledge that redundancy can be a key feature of effective storytelling, particularly for ensuring coherence (*81*). Therefore, we do not interpret higher LZ complexity as a direct measure of creativity or writing quality. Rather, we include it as a supplementary, information-theoretic lens that captures one limited aspect of text diversity.

*Statistical tests*

To evaluate the statistical significance of our findings, we employed two-sided independent samples t-tests to compare the distributions of responses under different conditions and accounted for multiple comparisons using the false discovery rate (FDR) correction. We also calculated effect sizes using Cohen's $d$ to quantify the magnitude of differences between groups (see Fig S3).

*PCA of text embeddings*

To assess variation in writing style and content between humans and different LLMs, we embedded the creative writings using a text embedding model and visualized a low-dimensional representation of the embeddings using Principal Component Analysis (PCA). For this analysis, we used the `text-embedding-ada-002` model from OpenAI's API to embed the entirety of the text into a single 1536-dimensional vector. We then applied PCA to the texts authored by humans and those generated by different LLMs. Distinct PCA models were used for each specific writing task, which included haikus, flash fiction, and synopses. Finally, we created visualizations of the distributions of the first two principal components to evaluate the similarity of the produced texts in the embedding space.

**Data Availability**

All data supporting the findings of this study are openly available. The dataset can be accessed on the Open Science Framework (OSF) at https://osf.io/z4c9a/?view_only=e9fb212880224572b780ace2f7102710. This ensures full transparency and reproducibility of the research. There are no restrictions on the materials used in this study, and all resources are detailed in the main text or supplementary materials.

## Code Availability

The code used for data acquisition, analysis, and visualization is openly available on GitHub at https://github.com/AntoineBellemare/DAT_GPT. Readers can access the full code repository without any restrictions to replicate or extend the analyses conducted in this study.

## Acknowledgments

### Funding:

A.B. was supported by Fonds de Recherche du Québec-Société et Culture doctoral grant (274043). K.J. was supported by Canada Research Chairs program funding (950-232368), Discovery Grant from the Natural Sciences and Engineering Research Council of Canada (2021-03426) and Strategic Research Clusters Program from the Fonds de recherche du Québec–Nature et technologies (2023-RS6-309472). J.O. was supported by Canadian

Institutes of Health Research postdoctoral fellowship. F.L. and Y.H. were supported by Courtois-Neuromod scholarships. At the time of submission, F.L. is supported by the Social Science and Humanities Research Council of Canada doctoral fellowship grant and the Applied AI Institute of Concordia University, Montreal.

**Author contributions:**

Author contributions were as follows: Conceptualization: ABP, FL, KJ; Methodology: ABP, FL, PT, YH, KJ; Investigation: ABP, FL, PT; Visualization: ABP, FL, PT; Supervision: JO, YB, KJ; Writing—original draft: ABP, FL, PT, YH, KJ; Writing—review & editing: ABP, FL, PT, YH, KM, JO, YB, KJ

**Competing interests:** K.W.M. is a Senior Research Scientist at Google DeepMind (GDM), but this work was conducted independently and was not part of their duties at GDM. All other authors declare no competing interests.

**Inclusion & Ethics :**
The protocol for human data collection was approved by the University of Toronto Research Ethics Board (#45872) and was deemed exempt by the Harvard University Institutional Review Board (IRB21-0991).

Supplementary Materials for

# Divergent Creativity in Humans and Large Language Models

*Bellemare-Pepin et al.*

*Corresponding author. Email: karim.jerbi@umontreal.ca

**This PDF file includes:**

**Supplementary Text**

Cosine similarity between haiku embeddings and the word “nature”.

The results depicted in Figure S1 demonstrate a clear trend: GPT-4-generated haikus consistently maintain a strong thematic focus on nature across all temperature settings, as evidenced by the generally high cosine similarity scores when compared to the word 'nature'. This observation suggests that GPT-4 adheres closely to the guideline of incorporating natural themes within the haikus it produces. This adherence could account for the non-significant effect of temperature variation on the Divergent Semantic Integration (DSI) scores observed in Figure 7. The relatively stable semantic alignment with the nature theme, regardless of the temperature changes, implies that GPT-4's haiku generation process is robust against such variations when it comes to maintaining thematic consistency.

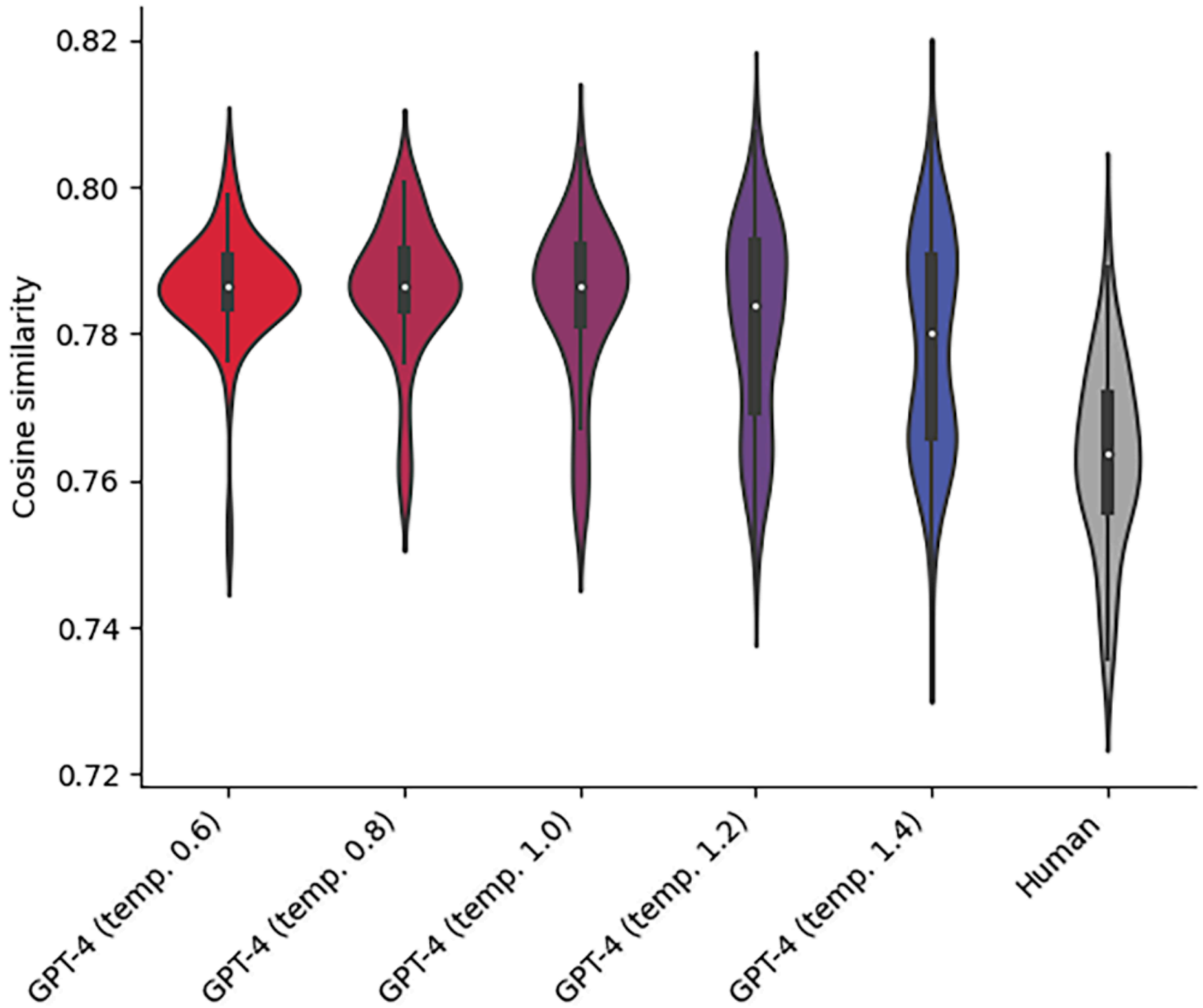

**Fig. S1. Cosine similarity between haiku embeddings and the word "nature".** The models represented include GPT-4 at different temperatures—0.6, 0.8, 1.0, 1.2, and 1.4—alongside a comparison with haikus authored by humans. Each plot shows the distribution of scores, with the white dot indicating the median value, the thick black bar showing the interquartile range, and the thin black lines denoting the overall range of data. The spread of each plot suggests the degree to which the haikus are semantically related to 'nature',

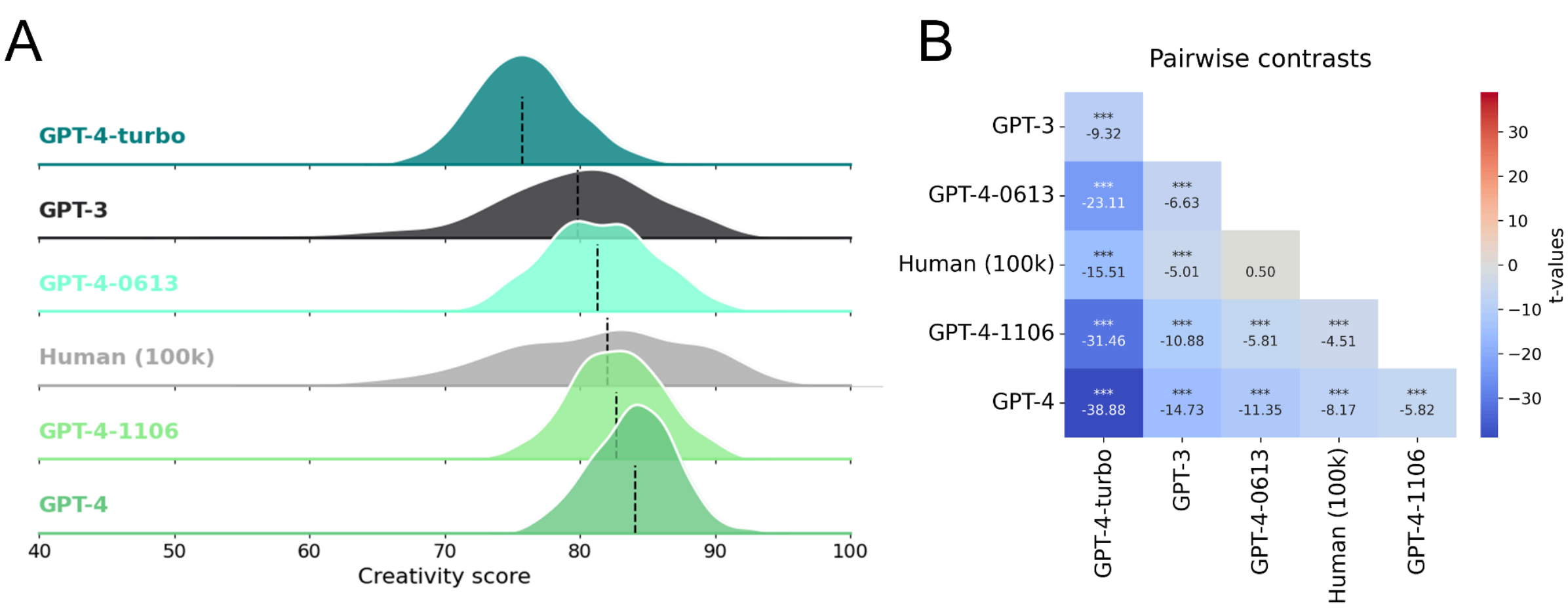

**Fig. S2. Comparing GPT-4 models and humans on the Divergent Association Task (DAT).** The models represented include GPT-3.5-turbo, GPT-4-0314, GPT-4-0613, GPT-4-1106 and GPT-4-turbo. **(A)** Mean DAT score and 95% confidence intervals. **(B)** Heatmap of all contrasts using independent t-tests, sorted by their correlation with the highest performing model, GPT-4-0314.

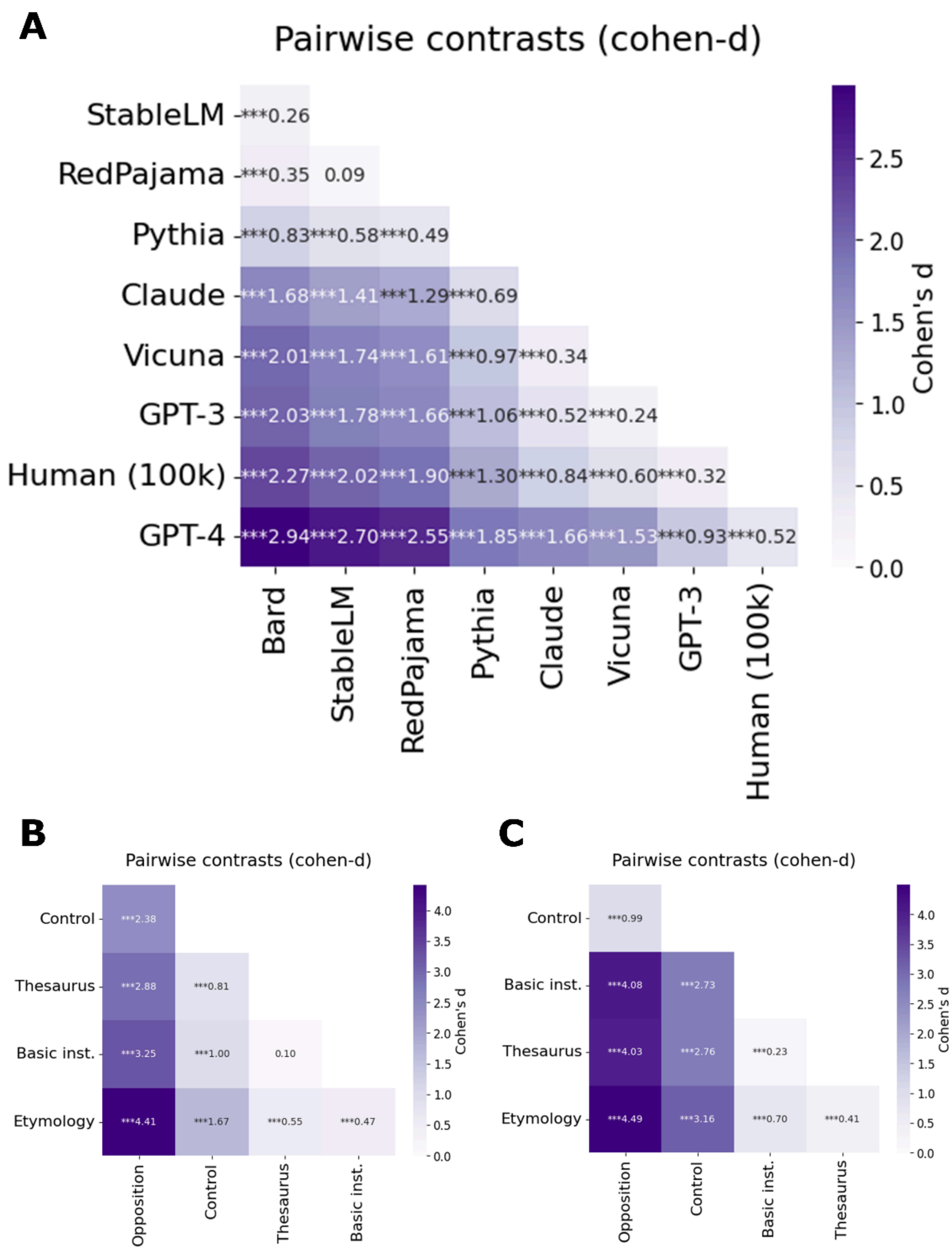

**Fig. S3. Effect size of the main results from Figures 1 and 4. (A)** Heatmap illustrating *Cohen's d* for all pairwise comparisons of DAT scores between LLMs and human responses. **(B)** Heatmap showing *Cohen's d* for comparisons of DAT scores across different GPT-3.5 strategies. **(C)** Heatmap depicting *Cohen's d* for comparisons of DAT scores across different GPT-4 strategies.

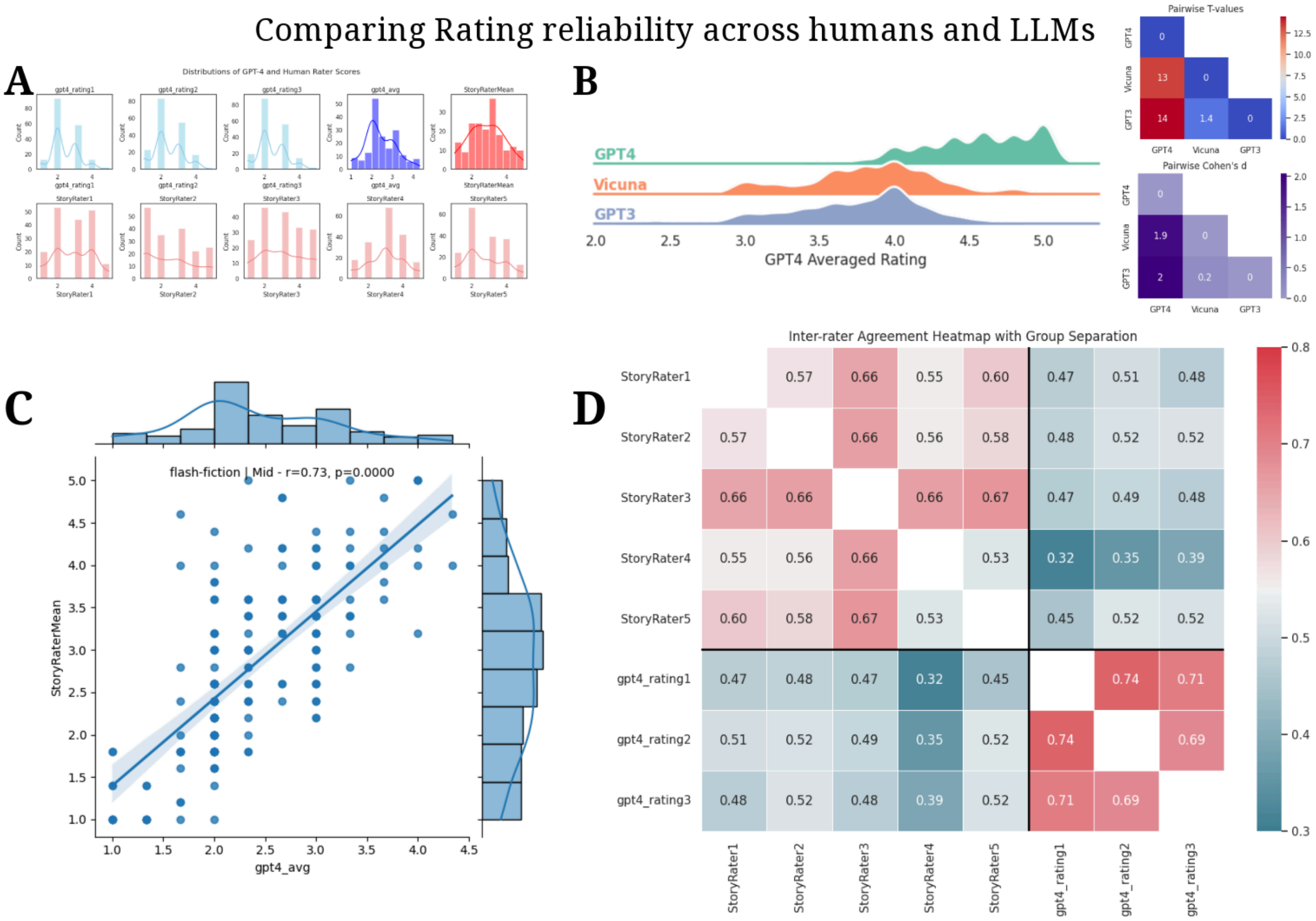

**Fig. S4. Assessment of inter-rater reliability across humans and GPT-4. (A)** Histograms showing the distribution of creativity scores given on an ordinal scale between 1 and 5 representing judgements of overall perceived creativity. **(B)** Distribution of GPT-4 averaged creativity judgements (ratings) show a significant difference between GPT-4 and Vicuna/GPT-3 [add stats]**(C)** A significant correlation was found between averaged judgements in GPT-4 and human raters [add stats] **(D)** Heatmap showing the inter-rater agreement of these judgements. There was a significantly lower concordance between human and GPT-4 raters ($M$ = 0.47, $SD$ = 0.06) compared to within human judgments ($M$ = 0.61, $SD$ = 0.05), $t(28) = 5.86$, $p < .001$, $d$ = 2.54.

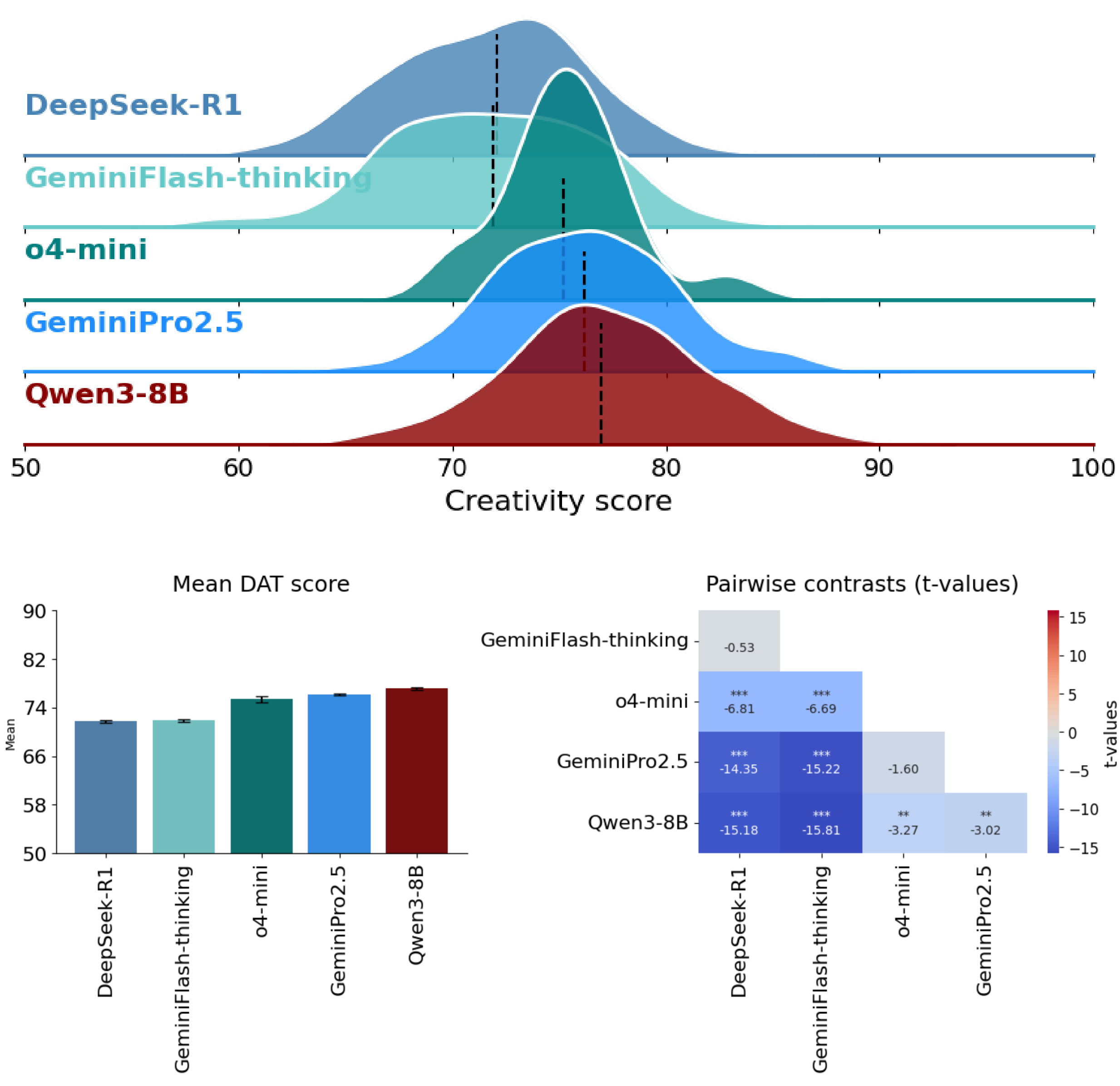

**Fig S5. Comparing "Reasoning" models on the Divergent Association Task (DAT).**

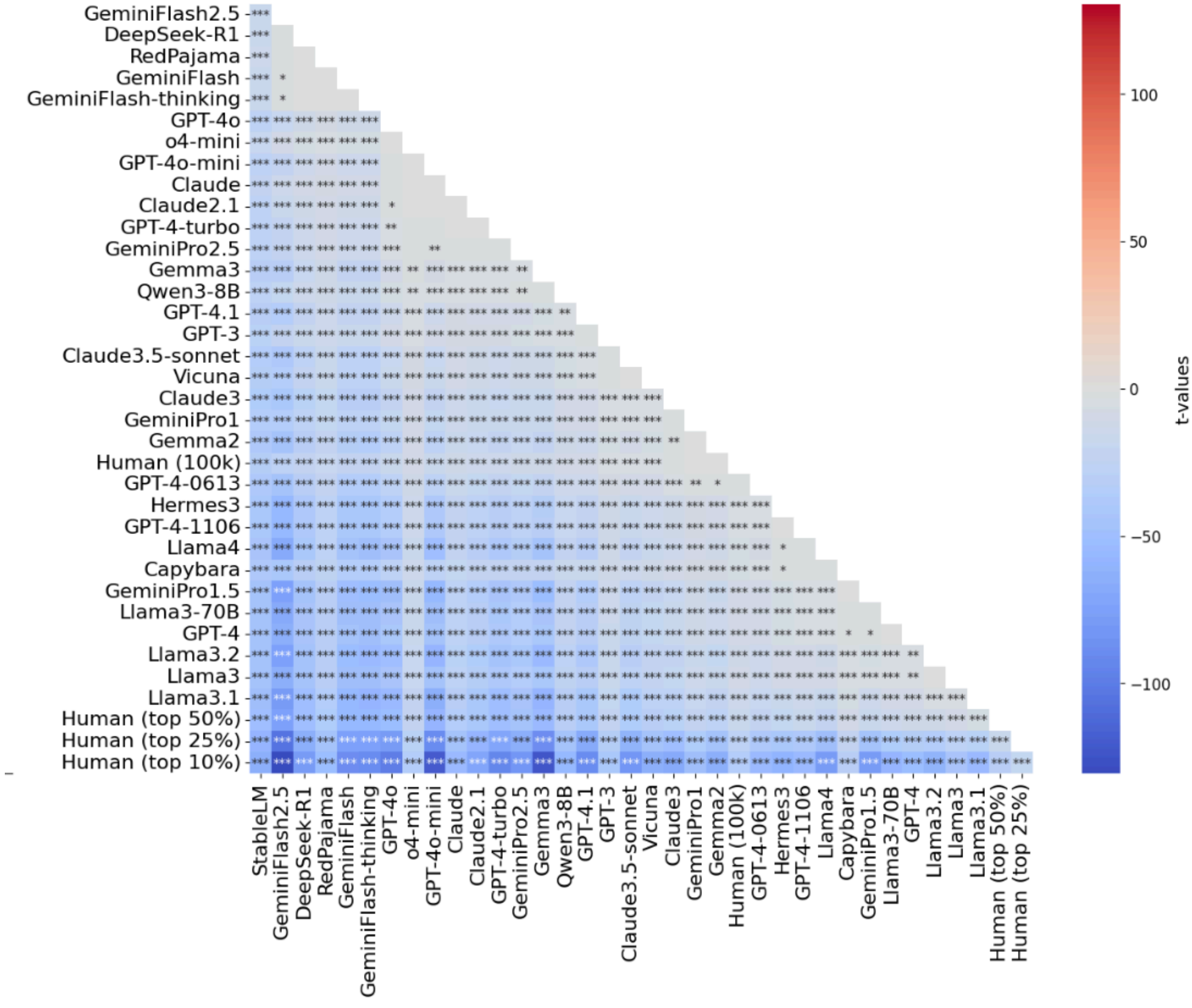

**Fig S6. Pairwise independent t-tests (t-values) comparing mean creativity scores across large language models (LLMs) and human samples on the Divergent Association Task (DAT).** Models and human benchmarks (full distribution, top 50%, 25%, and 10%) are ranked from lowest to highest mean score. Each cell displays the t-value for the contrast between the row and column models. Statistical significance is indicated by asterisks (* $p < .05$, ** $p < .01$, *** $p < .001$) after false discovery rate (FDR) correction for multiple comparisons.

| Model name | Organization | Model ID | Model size | Fine-tuning | Temp range/ default | License | Source | Year of release |
|---|---|---|---|---|---|---|---|---|
| Capybara | LAION | laion/oasst-sft-7b-llama-3.1 | 7B | RLHF | 0.7 | Apache 2.0 | Hugging Face | 2023 |
| Claude | Anthropic | claude-instant-1.2 | unknown | RLAIF | 0-1 / 0.7 | Paid access | Anthropic API | 2023 |
| Claude2.1 | Anthropic | claude-2.1 | unknown | RLAIF | 0-1 / 0.7 | Paid access | Anthropic API | 2023 |
| Claude3 | Anthropic | claude-3-opus-20240229 | unknown | RLAIF | 0-1 / 0.7 | Paid access | Anthropic API | 2024 |
| Claude3.5-sonnet | Anthropic | claude-3-5-sonnet-20240620 | unknown | RLAIF | 0-1 / 0.7 | Paid access | Anthropic API | 2024 |
| DeepSeek-R1 | DeepSeek AI | deepseek-coder-6.7b-instruct | 6.7B | RLHF | 0.7 | MIT | Hugging Face | 2024 |
| Gemma2 | Google | gemma-2-9b-it | 9B | Instruction Tuning | 0.7 | Gemma License | Google AI | 2024 |
| Gemma3 | Google | gemma3:27b | 27B | RLHF | n.a. | Gemma 3 License | ollama | 2025 |
| GeminiFlash | Google | gemini-1.5-flash-latest | 1.5M tokens | Instruction Tuning | 0-2 / 1.0 | Paid access | Google AI Platform | 2024 |
| GeminiFlash-thinking | Google | gemini-2.0-flash-thinking-exp-1219 | Undisclosed | Undisclosed | n.a. | Paid access | Google AI | 2024 |
| GeminiPro1 | Google | gemini-1.0-pro | unknown | Instruction Tuning | 0-1 / 0.9 | Paid access | Google AI Platform | 2023 |
| GeminiPro1.5 | Google | gemini-1.5-pro-latest | 1M tokens | Instruction Tuning | 0-2 / 1.0 | Paid access | Google AI Platform | 2024 |
| GeminiPro2.5 | Google | gemini-2.5-pro-preview-05-06 | Undisclosed | Undisclosed | n.a. | Paid access | Google AI | 2025 |
| GeminiFlash2.5 | Google | gemini-2.5-flash | Undisclosed | Undisclosed | n.a. | Paid access | Google AI | 2025 |
| GPT-3 | OpenAI | gpt-3.5-turbo | 175B | RLHF | 0-2 / 1 | Paid access | OpenAI API | 2022 |
| GPT-4 | OpenAI | gpt-4 | unknown | RLHF | 0-2 / 1 | Paid access | OpenAI API | 2023 |
| GPT-4-0613 | OpenAI | gpt-4-0613 | unknown | RLHF | 0-2 / 1 | Paid access | OpenAI API | 2023 |
| GPT-4-1106 | OpenAI | gpt-4-1106-preview | 128k tokens | RLHF | 0-2 / 1 | Paid access | OpenAI API | 2023 |
| GPT-4-turbo | OpenAI | gpt-4-turbo | 128k tokens | RLHF | 0-2 / 1 | Paid access | OpenAI API | 2024 |
| GPT-4o | OpenAI | gpt-4o | unknown | RLHF | 0-2 / 1 | Paid access | OpenAI API | 2024 |
| GPT-4o-mini | OpenAI | gpt-4o-mini | unknown | RLHF | 0-2 / 1 | Paid access | OpenAI API | 2024 |
| GPT-4.1 | OpenAI | gpt-4.1-2025-04-16 | Undisclosed | Undisclosed | 0.0 - 2.0 /1.0 | Paid access | OpenAI API | 2025 |
| Hermes3 | Nous Research | Nous-Hermes-2-Yi-34B | 34B | Instruction Tuning | 0.7 | Apache 2.0 | Hugging Face | 2023 |
| Llama3 | Meta | meta-llama/Llama-3-8B-Instruct | 8B | RLHF | 1 | Llama 3 License | Hugging Face | 2024 |
| Llama3-70B | Meta | meta-llama/Llama-3-70B-Instruct | 70B | RLHF | 1 | Llama 3 License | Hugging Face | 2024 |
| Llama3.1 | Meta | meta-llama/Llama-3.1-8B-Instruct | 8B | RLHF | 1 | Llama 3.1 License | Hugging Face | 2024 |
| Llama3.2 | Meta | llama3.2:3b | 3B | RLHF | 1 | Llama 3.2 Community License Agreement | ollama | 2024 |
| Llama4 | Meta | llama4:latest | 16x17B | RLHF | 1 | Llama 4 Community License Agreement | ollama | 2025 |
| o4-mini | OpenAI | o4-mini-2025-04-16 | unknown | RLHF | 1 | Paid access | OpenAI API | 2025 |
| Qwen3-8B | Alibaba Cloud | Qwen1.5-7B-Chat | 7B | RLHF | n.a. | Apache 2.0 | Hugging Face | 2024 |
| RedPajama | Together | RedPajama-INCITE-Chat-7B-v0.1 | 7B | RLHF | 0-1 / 0.7 | Apache 2.0 | Hugging Face | 2023 |
| StableLM | Stability AI | stablelm-2-zephyr-1.6b | 1.6B | RLHF | n.a. | Stability AI License | Hugging Face | 2024 |
| Vicuna | LMSYS | lmsys/vicuna-7b-v1.5 | 7B | RLHF | 0.7 | Apache 2.0 | Hugging Face | 2023 |

**Table S1. Full list of model specifications.**